\documentclass{article}

    \usepackage[final]{ewrl_2024}

\usepackage[utf8]{inputenc} %
\usepackage[T1]{fontenc}    %
\usepackage[backref=page]{hyperref}       %
\usepackage{url}            %
\usepackage{booktabs}       %
\usepackage{amsfonts}       %
\usepackage{nicefrac}       %
\usepackage{microtype}      %
\usepackage{xcolor}         %
\usepackage{amsmath}
\usepackage[pdftex]{graphicx}
\usepackage{comment}
\usepackage{caption}
\usepackage{subcaption}
\usepackage[ruled]{algorithm2e}
\usepackage{algpseudocode}
\usepackage{wrapfig}

\usepackage{amsthm}
\newtheorem{definition}{Definition}

\definecolor{highlight}{RGB}{153, 0, 0}
\definecolor{highlight}{rgb}{0.9, 0, 0}

\title{Explore-Go: Leveraging Exploration for Generalisation in Deep Reinforcement Learning}

\author{%
  Max Weltevrede \\
  Department of Intelligent Systems \\
  Delft University of Technology \\
  Delft, 2628 XE, The Netherlands \\
  \texttt{m.r.weltevrede@tudelft.nl}
  \And
  Felix Kaubek \\
  Department of Intelligent Systems \\
  Delft University of Technology \\
  Delft, 2628 XE, The Netherlands 
   \AND
   Matthijs T. J. Spaan \\
  Department of Intelligent Systems \\
  Delft University of Technology \\
  Delft, 2628 XE, The Netherlands 
   \And
   Wendelin B\"ohmer \\
  Department of Intelligent Systems \\
  Delft University of Technology \\
  Delft, 2628 XE, The Netherlands 
}

\begin{document}

\maketitle

\begin{abstract}
    One of the remaining challenges in reinforcement learning is to develop agents that can generalise to novel scenarios they might encounter once deployed. This challenge is often framed in a multi-task setting where agents train on a fixed set of tasks and have to generalise to new tasks. Recent work has shown that in this setting increased exploration during training can be leveraged to increase the generalisation performance of the agent. This makes sense when the states encountered during testing can actually be explored during training. In this paper, we provide intuition why exploration can also benefit generalisation to states that cannot be explicitly encountered during training. Additionally, we propose a novel method \emph{Explore-Go} that exploits this intuition by increasing the number of states on which the agent trains. Explore-Go effectively increases the starting state distribution of the agent and as a result can be used in conjunction with most existing on-policy or off-policy reinforcement learning algorithms. We show empirically that our method can increase generalisation performance in an illustrative environment and on the Procgen benchmark. 
\end{abstract}

\section{Introduction}
Despite the advances in reinforcement learning (RL) in recent years, it is still rare for RL to be applied to real-world problems. One of the remaining challenges is that an agent deployed in the real world needs the ability to generalise to novel scenarios it might encounter. This is the main research question explored in the zero-shot policy transfer (ZSPT) setting \citep{kirk_survey_2023}. In the ZSPT setting, the agent gets to train on several tasks (instances of an environment) and needs to generalise to new ones. This differs from the traditional single-task RL setting, in which the agent trains and tests on the same environment instance. 

Aside from generalisation, another important challenge in RL is that of the exploration-exploitation trade-off. The trade-off is characterised by the decision of how much an agent should explore new trajectories, versus how much it should exploit trajectories that are known to be good at that time. In single-task RL, the optimal trade-off usually comes down to exploring just enough to be able to solve the training task optimally. Recent work \citep{jiang_importance_2023, weltevrede_role_2023} has re-evaluated this trade-off for the ZSPT setting or has more generally asked the question: what data should we ideally train on to generalise best to new tasks? Both works demonstrate that skewing the exploration-exploitation trade-off more towards exploration can improve generalisation performance. 

\begin{wrapfigure}{r}{0.35\textwidth}
\vspace{-0mm}
\centering
\includegraphics[width=.34\textwidth]{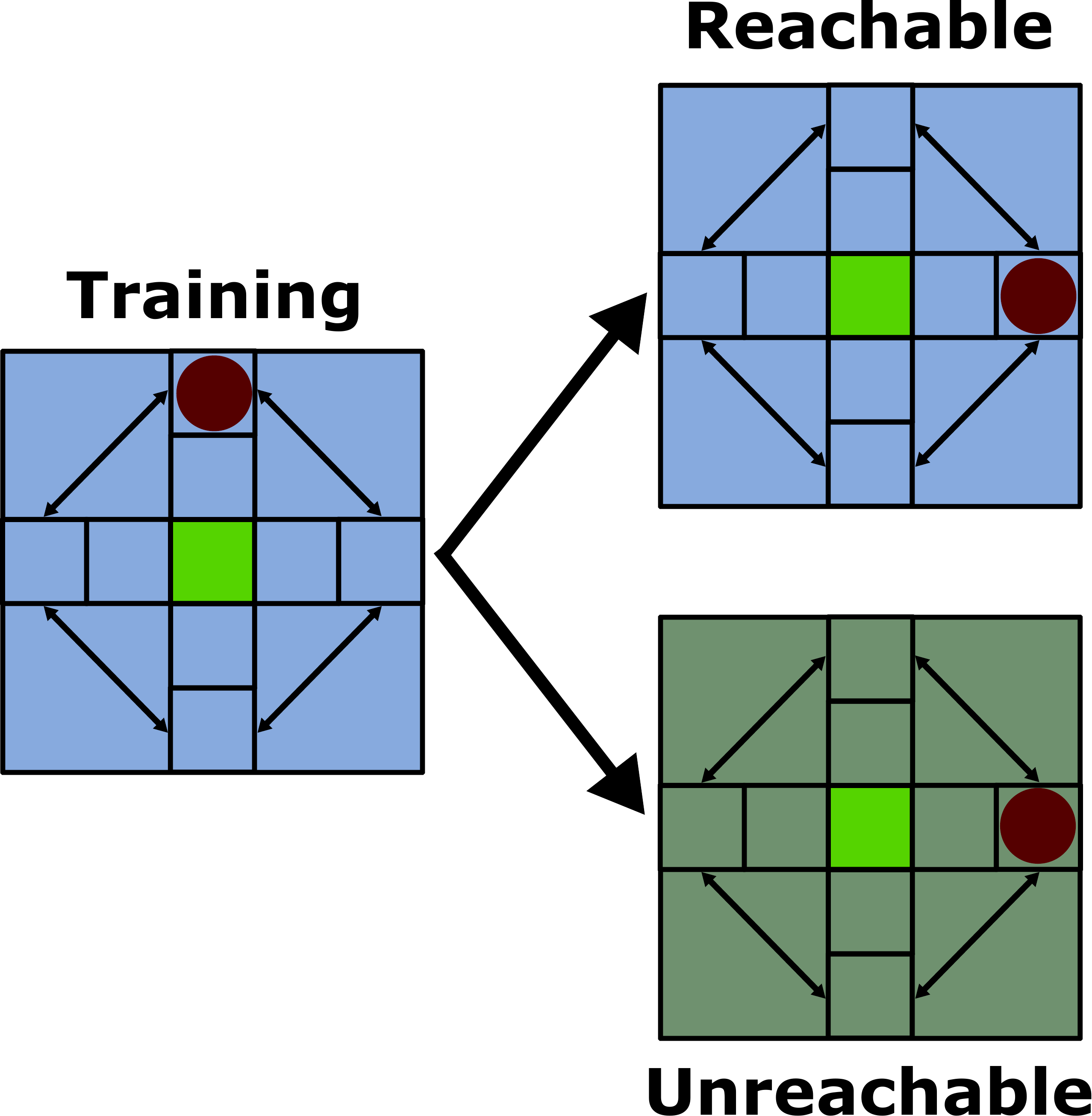}
\caption{Example of a reachable and unreachable task. The agent (circle) needs to move to the goal location (light green square). 
The reachable task on the right has a different start state, which can be reached from the training task. The unreachable task differs by the background and cannot be reached.}
\label{fig:reachable}
\vspace{-0.5mm}
\end{wrapfigure}

\citet{weltevrede_role_2023} propose a distinction between the types of tasks we want to generalise to:  reachable vs unreachable. Reachable tasks share states and rewards with the training tasks and can therefore be explored during training, whereas unreachable tasks cannot (see Figure \ref{fig:reachable} for an example). 
They argue this distinction is useful as the states encountered in reachable tasks, contrary to the states in unreachable tasks, can be optimised for during training. Therefore, exploring and learning to solve as many reachable states as possible, even if they are not necessary to solve the initial training tasks, will logically lead to improved reachable generalisation performance \citep{jiang_importance_2023, weltevrede_role_2023}. The same logic does not directly apply to generalising to unreachable tasks. For example, an agent can learn to react correctly to all reachable states, and therefore all reachable tasks, but since unreachable tasks contain states that cannot been seen during training, generalisation to unreachable tasks can be arbitrarily bad. Despite this, both \citet{jiang_importance_2023} and \citet{weltevrede_role_2023} show empirically that exploring the reachable state space improves generalisation to unreachable tasks. However, they do not provide intuition for why this is the case.\footnote{\citet{jiang_importance_2023} do not explicitly make the distinction between reachable and unreachable generalisation,  but we argue in Appendix \ref{app:related} that their intuition primarily applies to the reachable generalisation setting.}

The goal of this paper is to provide this intuition for why exploring and training on more of the reachable state space can also benefit generalisation to unreachable tasks. Our contributions are the following:
\begin{itemize}
    \item We introduce intuition on how unreachable generalisation can be improved by viewing training on more reachable states as a form of implicit data augmentation.
    \item We propose a novel method \emph{Explore-Go} that can be combined with most existing RL algorithms. Explore-Go performs exploration at the start of every episode in order to train on more of the reachable state space.  We also verify empirically that this method can improve generalisation to unreachable tasks. 
\end{itemize}

\section{Background}
A Markov decision process (MDP) $\mathcal{M}$ is a sequential decision making process defined by a 6-tuple $\mathcal{M} = \{S, A, R, T, p_0, \gamma \}$. In this definition, $S$ denotes a set of states called the state space, $A$ a set of actions called the action space, $R: S \times A \to \mathbb{R}$ the reward function, $T: S \times A \to \mathcal{P}(S)$ the transition function where $\mathcal{P}(S)$ denotes the set of probability distributions over states $S$, $p_0: \mathcal{P}(S)$ the starting state distribution and $\gamma \in [0,1)$ a discount factor. The goal is to find a policy $\pi: S \to \mathcal{P}(A)$ that maps states to probability distributions over actions in such a way that maximises the expected cumulative discounted reward $\mathbb{E}_{\pi} [ \sum_{t=0}^\infty \gamma^t r_t ]$, also called the \emph{return}. The expectation $\mathbb{E}_{\pi}$ is over the Markov chain $\{s_0, a_0, r_0, s_1, a_1, r_1 ... \}$ induced by policy $\pi$ when acting in MDP $\mathcal{M}$ \citep{akshay_steady-state_2013}. An optimal policy $\pi^*$ is one that achieves the highest possible return. The on-policy distribution $\rho^\pi : \mathcal{P}(S)$ of the Markov chain induced by policy $\pi$ in MDP $\mathcal{M}$ defines the proportion of time spent in each state as the number of episodes in $\mathcal{M}$ goes to infinity \citep{sutton_reinforcement_2018}.

\subsection{Contextual Markov decision process}
A contextual MDP \citep[CMDP,][]{hallak_contextual_2015} is a specific type of MDP where the state space $S = S' \times C$ can in principle be factored into an underlying state space $S'$ and a context space $C$. For a state $s = (s',c) \in S$, the context $c$ behaves differently than the underlying state $s'$ in that it is sampled at the start of an episode (as part of the distribution $p_0$) and remains fixed until the episode ends. The context $c$ can be thought of as the task an agent has to solve and from here on out we will refer to context and task interchangeably. 

The zero-shot policy transfer \citep[ZSPT,][]{kirk_survey_2023} setting for CMDPs $\mathcal{M}|_{C}$ considered in this paper is defined by a distribution over task space $\mathcal{P}(C)$ and a set of tasks $C^{train}$ and $C^{test}$ sampled from the same distribution $\mathcal{P}(C)$. The goal of the agent is to maximise performance in the testing CMDP $\mathcal{M}|_{C^{test}}$ defined by the CMDP induced by the testing tasks $C^{test}$, but is only allowed to train in the training CMDP $\mathcal{M}|_{C^{train}}$. The agent is expected to perform \emph{zero-shot} generalisation for the testing tasks, without any fine-tuning or adaptation period. 

In general, the task $c$ can influence several aspects of the underlying MDP, like the reward function or dynamics of the environment. As a result,  several existing fields of study like multi-goal RL (task influences reward) or sim-to-real transfer (task influences dynamics and/or visual observations) can be framed as special instances of the CMDP framework. However, in this paper we consider the specific CMDP setting where the task $c$ only influences the starting state distribution $p_0$. This means the only difference between tasks is their starting state $s'_0$. This is a common setting for generalisation research in reinforcement learning that describes several environments from the popular Procgen and Minigrid benchmarks \citep{cobbe_leveraging_2020, chevalier-boisvert_minigrid_2023}. 

\subsubsection{Reachability}
In this setting, tasks may start in different states but can still \emph{share} states $s'_t$ later in the episode. For example, if tasks have different starting positions but share the same goal, or if the agent can manipulate the environment to resemble a different task. Some tasks are \emph{unreachable} though. An example would be a task with a completely new background colour, as shown in Figure \ref{fig:reachable}: no action the agent performs can change the background in that episode. In this setting, we can refer to tasks $c$ and underlying states $s' \in S'$ interchangeably since we can think of any $s'$ as a starting state and therefore as a task. As a result, we will simplify notation by dropping the apostrophe and referring to underlying states $s' \in S'$ as states $s \in S$ and tasks $c \in C$ as starting states $s_0 \in S_0$.  

Formally, we define reachability of states in the CMDP $\mathcal{M}|_{S_0^{train}}$ as in \citep{weltevrede_role_2023}:
\begin{definition}
\label{thrm:reachability}
    The set of reachable states $S_r(\mathcal{M}|_{S_0^{train}})$ (abbreviated with $S_r$ from now on) consists of all states $s_r$ for which there exists a sequence of actions that give a non-zero probability of ending up in $s_r$ when performed in the CMDP $\mathcal{M}|_{S_0^{train}}$.
\end{definition}
Put differently, a state $s_r$ is reachable if there exists a policy whose probability of encountering that state during training is non-zero and the set of all reachable states is denoted with $S_r$.  In complement to reachable states, we define \emph{unreachable} states $s_u$ as states that are not reachable.

We can now define two instances of the ZSPT problem: ZSPT to reachable states and ZSPT to unreachable states. ZSPT to reachable states, which we will refer to as the \emph{reachable generalisation} setting, is a ZSPT problem where the initial states during testing $S_0^{test}$ are part of the set of reachable states during training $S_0^{test} \subseteq S_r$. Due to how reachability is defined, in the reachable generalisation setting all states encountered in the testing CMDP $\mathcal{M}|_{S_0^{test}}$ are also reachable. Note that the reverse does not have to be true: not all reachable states can necessarily be encountered in $\mathcal{M}|_{S_0^{test}}$. 

Correspondingly, the ZSPT to unreachable states, which we refer to as the \emph{unreachable generalisation} setting, is a ZSPT problem where the initial states during testing $S_0^{test}$ are \textbf{not} part of the set of reachable states during training $S_0^{test} \cap S_r = \emptyset$. We will assume in the unreachable generalisation setting all states encountered in the testing CMDP $\mathcal{M}|_{S_0^{test}}$ are also unreachable.\footnote{This holds for ergodic CMDPs. However, in some non-ergodic CMDPs it is possible that you can transition into the reachable set $S_r$ after starting in an unreachable state, which we won't consider in this paper.} Note that even though the starting states during testing are unreachable, it is still considered \emph{in-distribution} generalisation since they are sampled from the same distribution as the starting states during training.

\subsection{Exploration \& experience replay for generalisation}
\label{sec:background-expl}
In the single-task setting, the goal is to maximise performance in the MDP $\mathcal{M}$ in which the agent trains. In this setting, it is sufficient to learn an optimal policy in all the states $s \in S$ encountered by the optimal policy in $\mathcal{M}$. This is because acting optimally in all the states encountered by the optimal policy in $\mathcal{M}$ guarantees maximal return in $\mathcal{M}$. This means that exploration and experience replay only have to facilitate learning the optimal policy on the on-policy distribution $\rho^{\pi^*}(\mathcal{M})$. In fact, once the optimal policy has been found, learning to be optimal anywhere else in $\mathcal{M}$ would be a wasted effort that potentially allocates approximation power to unimportant areas of the state space.  

Recent work has shown that this logic does not transfer to the ZSPT problem setting. In this setting, the goal is not to maximise performance in the training CMDP $\mathcal{M}|_{S_0^{train}}$, but rather to maximise performance in the testing CMDP $\mathcal{M}|_{S_0^{test}}$. Ideally, the learned policy will be optimal over the on-policy distribution $\rho^{\pi^*}(\mathcal{M}|_{S_0^{test}})$ in this testing CMDP. However, in general this distribution is unknown. Instead, \citet{jiang_importance_2023, weltevrede_role_2023} suggest the next best thing is to learn a policy that is optimal over as much of the reachable state space $S_r$ as possible. 

In the reachable generalisation setting, the states in $\rho^{\pi^*}(\mathcal{M}|_{S_0^{test}})$ are (by definition) part of the reachable state space $S_r$. So, through more extensive exploration in the training environments, an agent can gather more knowledge that can be used to generalise to new environments \citep{jiang_importance_2023}. Moreover, if a policy were optimal on all reachable states, it would be guaranteed to 'generalise' to reachable tasks \citep{weltevrede_role_2023}. One could argue generalisation is not the best term to use here, since even a policy that completely overfits to the reachable state space $S_r$ would exhibit perfect 'generalisation'.

\section{Unreachable generalisation}
\label{sec:unreachable}
\begin{figure}[ht]
\begin{center}
    \begin{subfigure}[b]{0.49\textwidth}
        \centering
        \includegraphics[width=1\textwidth]{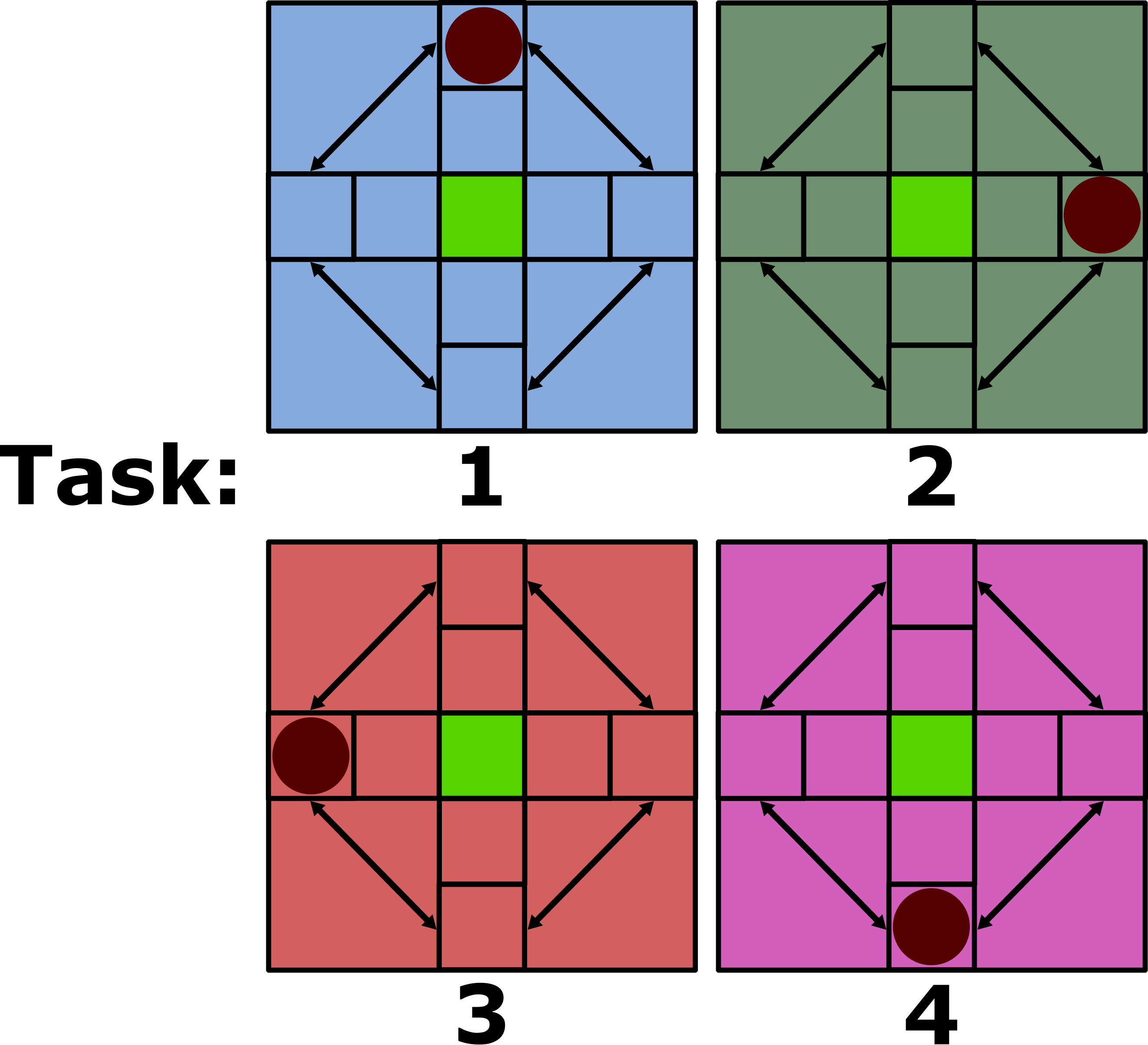}
        \caption{Illustrative CMDP}
        \label{fig:illustrative-tasks}
    \end{subfigure}
    \hfill
    \begin{subfigure}[b]{0.49\textwidth}
        \centering
        \includegraphics[width=1\textwidth]{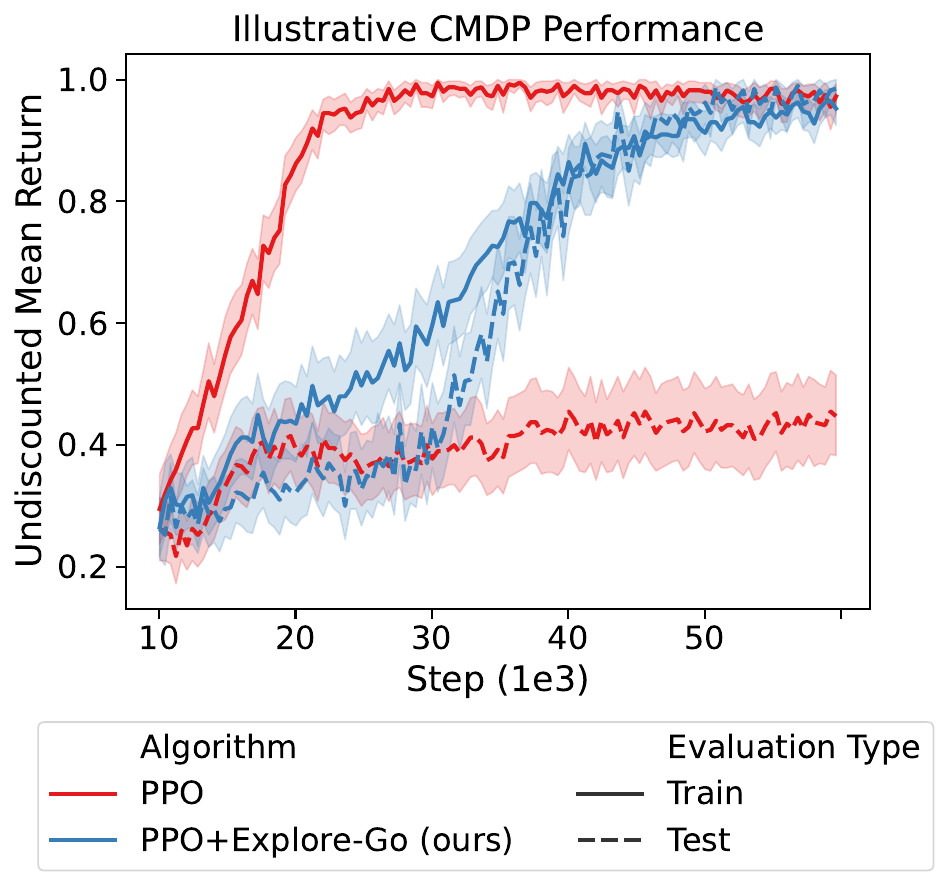}
        \caption{Performance of PPO and PPO+Explore-Go (ours).}
        \label{fig:illustrative-results}
    \end{subfigure}
    \newline
    \begin{subfigure}[b]{0.49\textwidth}
        \centering
        \includegraphics[width=1\textwidth]{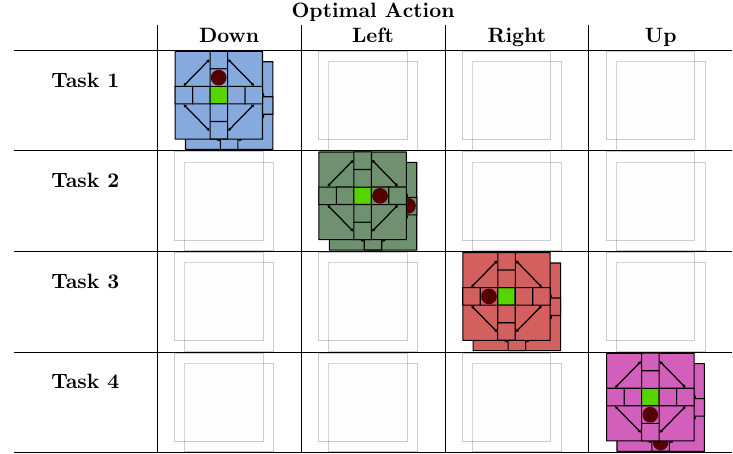}
        \caption{States along optimal trajectory}
        \label{fig:illustrative-optimal}
    \end{subfigure}
    \hfill
    \begin{subfigure}[b]{0.49\textwidth}
        \centering
        \includegraphics[width=1\textwidth]{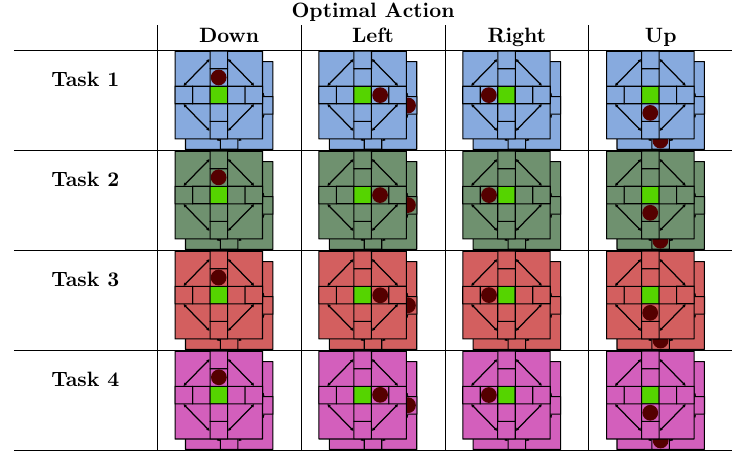}
        \caption{Full state space}
        \label{fig:illustrative-full}
    \end{subfigure}
\end{center}
\caption{(a) Illustrative CMDP with four training tasks, each differing in background colour and agent (circle) starting position. All tasks share the same goal location (green square in the middle). (b) Performance of a baseline PPO agent and our Explore-Go agent on the CMDP. The agent trains on the tasks in (a) and is tested in tasks with a completely new background colour. Shown are mean and 95\% confidence interval over 100 seeds. Below are (c) the states along the optimal trajectories, (d) the reachable state space, categorised by the task they're from (rows) and the optimal action (columns).}
\label{fig:illustrative}
\end{figure}

For unreachable generalisation, the states encountered in $\rho^{\pi^*}(\mathcal{M}|_{S_0^{test}})$ are not part of the reachable space $S_r$. Therefore, it is not immediately obvious over what part of the state space we should be optimal. However, in this section we will argue that in the unreachable generalisation setting, much like for reachable generalisation, the learned policy should be optimal over as much of the reachable state space $S_r$ as possible. 

To illustrate this, we define an illustrative CMDP in Figure \ref{fig:illustrative-tasks}. This CMDP consists of a cross-shaped grid world with additional transitions that directly move the agent between adjacent end-points of the cross (e.g., moving right at the end-point of the northern arm of the cross will move you to the eastern arm). The goal for the agent (circle) is to move to the centre of the cross (the green square). There are four different training tasks which differ in the starting location of the agent and the colour of the background. 

Let us first consider the states on which an agent has to be optimal in the single-task RL setting: the states along the optimal trajectories. In Figure \ref{fig:illustrative-optimal} these states are split into a table according to what task they are from (rows) and what action is optimal (columns). Formally, the columns can be defined through a state abstraction $\phi(s)$, where two states $s,s' \in S$ in a particular column map onto the same abstracted state $\phi(s) = \phi(s')$. Since we are learning policies in our example, we use the $\pi^*$-irrelevance state abstraction $\phi_{\pi^*}(s)$\citep{li_towards_2006}, which is defined such that for any states $s, s' \in S$, $\phi_{\pi^*}(s) = \phi_{\pi^*}(s')$ implies $\pi^*(s) = \pi^*(s')$. Now, along the optimal trajectories the colour of the background is perfectly correlated with the abstracted state $\phi_{\pi^*}(s)$. Therefore it is also correlated with what action is optimal. A policy trained to be optimal on only these states has a high likelihood of overfitting to this correlation. As a result, this policy is unlikely to generalise to new reachable states (missing cells in Figure \ref{fig:illustrative-optimal}), let alone to new unreachable states with an altogether different background colour (a completely new row).  This can be seen in Figure \ref{fig:illustrative-results} where a PPO agent (red) that mainly trains on these states does not generalise to tasks with a new background colour (see Section \ref{sec:exp-ill} for more on this experiment). 

Suppose now, we have a policy that is optimal over the entire reachable state space (see Figure \ref{fig:illustrative-full}). This policy has likely learned to ignore the colour of the background, as on the entire reachable state space this no longer correlates with what action to take. As such, this policy is now more likely to generalise to new tasks with different background colours. We see this in our novel method PPO+Explore-Go (blue in Figure \ref{fig:illustrative-results}), which trains on all reachable states, and generalises to unseen background colours. One perspective on this is that learning a policy over more of the reachable state space reduces the probability of abusing correlations that may exist during training, but that do not exist during testing. In other words, it reduces the probability of overfitting to spurious correlations (between background colour and optimal action). Another perspective is that exploring and training on more states encourages the policy to become invariant to symmetries in state and task space (colour symmetry). 

More generally, we could think of the inclusion of the other reachable states in Figure \ref{fig:illustrative-full} (compared to Figure \ref{fig:illustrative-optimal}) as a form of data augmentation. For example, the additional states from tasks 2, 3 and 4 in the first column in Figure \ref{fig:illustrative-full}, can be viewed as being generated by a set of transformations that change the background colour but leave the function output (what action to take) invariant. Data augmentation techniques are commonly used to improve generalisation performance in a wide variety of settings and applications \citep{shorten_survey_2019, feng_survey_2021, zhang_understanding_2021, miao_learning_2023} and are thought to work by reducing overfitting to spurious correlations \citep{shen_data_2022}, inducing model invariance \citep{lyle_benefits_2020, chen_group-theoretic_2020} and/or regularising training \citep{bishop_training_1995, lin_good_2022}. 

Based on this perspective, we postulate that unreachable generalisation specifically benefits from training on more states belonging to the same abstracted state (defined by $\phi_{\pi^*}$ in this example). For instance, training on more states where the optimal action is to go down (states in the first column in Figure \ref{fig:illustrative-full}) will increase our likelihood of learning a function that is invariant to any differences between those states (like the background colour). This in turn will increase our likelihood of generalising correctly to a state with a completely new background colour, even if that state is not reachable during training. Of course, one could still learn a function that wrongly generalises to unseen test states, but without any additional knowledge of the structure of the testing tasks, we opt to simply train on as many reachable states as possible.

\section{Method}
\begin{wrapfigure}{R}{0.6\textwidth}
\vspace{-5mm}
\centering
\begin{algorithm}[H]
\SetKwIF{If}{ElseIf}{Else}{\textcolor{highlight}{if}}{\textcolor{highlight}{then}}{\textcolor{black}{else if}}{\textcolor{black}{else}}{\textcolor{black}{end if}}%

\caption{PPO + \textcolor{highlight}{Explore-Go}}
\label{alg:explore-go}
\KwIn{PPO agent $PPO$, \textcolor{highlight}{pure exploration agent} \textcolor{highlight}{$PE$}, \textcolor{highlight}{max number of pure exploration steps} \textcolor{highlight}{$K$}}
\textcolor{highlight}{$k \gets \mathit{Uniform}(0, K)$}\;
$i \gets 0$ \Comment{Counts steps within an episode}\;
\For{$iteration = 0,1,2,...$}{
    $\mathcal{D}_{PPO} \gets \{ \}$\;
    \textcolor{highlight}{$\mathcal{D}_{PE} \gets \{ \}$}\;
    \For{$step = 0,1,2,..., T$}{
        \eIf{\textcolor{highlight}{$i < k$}}{
            \textcolor{highlight}{Sample transition $t$ by running $PE$}\;
            \textcolor{highlight}{Add $t$ to $\mathcal{D}_{PE}$}\;
        }{
            Sample transition $t$ by running $PPO$\;
            Add $t$ to $\mathcal{D}_{PPO}$\;
        }
        $i \gets i + 1$\;
        
        \SetKwIF{If}{ElseIf}{Else}{\textcolor{black}{if}}{\textcolor{black}{then}}{\textcolor{black}{else if}}{\textcolor{black}{else}}{\textcolor{black}{end if}}%
        
        \If{end of episode}{
            \textcolor{highlight}{$k \gets \mathit{Uniform}(0, K)$}\;
            $i \gets 0$\;
            Reset environment\;
        }
    }

    Update $PPO$ with trajectories $\mathcal{D}_{PPO}$\;
    \textcolor{highlight}{(Optional) Update} \textcolor{highlight}{$PE$} \textcolor{highlight}{with trajectories} \textcolor{highlight}{$\mathcal{D}_{PE}$}\;
}
\end{algorithm}
\vspace{-4mm}
\end{wrapfigure}

In this section, we will propose a novel method that increases the distribution over states on which our agents trains with the goal of improving unreachable generalisation performance. Our method \emph{Explore-Go}\footnote{The name Explore-Go is a variation of the popular exploration approach Go-Explore \citep{ecoffet_first_2021}. In Go-Explore the agent at the start of every episode first teleports to a novel state and then continuous exploration. In our approach, the agent first explores until it finds a novel state and then goes and solves the original task.} can be combined with most existing RL algorithms. Our algorithm of choice is proximal policy optimisation \citep[PPO,][]{schulman_proximal_2017} as many approaches designed to improve zero-shot generalisation in contextual MDPs are based on it \citep{cobbe_phasic_2021, jiang_prioritized_2021, raileanu_decoupling_2021, moon_rethinking_2022}. PPO is an algorithm that requires (primarily) on-policy data for its training, distributed along the on-policy state distribution $\rho^{\pi_\theta}(\mathcal{M}|_{S_0^{train}})$ of the current policy $\pi_\theta$. Therefore, we cannot arbitrarily change the distribution of states over which our agent trains. However, the on-policy distribution depends on the starting states $S_0^{train}$, as where the agent starts influences what states it is likely to encounter. Therefore, one way to increase the coverage of the on-policy state distribution $\rho^{\pi_\theta}(\mathcal{M}|_{S_0^{train}})$ is to artificially increase the number of starting states $S_0^{train}$. 

Our method Explore-Go works by effectively increasing the diversity of the starting state distribution by performing a \emph{pure exploration phase} for a certain number of steps at the beginning of each episode (see Algorithm \ref{alg:explore-go} where red highlights the modifications Explore-Go makes to PPO). Pure exploration refers to an objective that ignores the rewards $r_t$ the agent encounters and instead focuses purely on exploring new parts of the state space. After the pure exploration phase, the state the agent has ended up in is treated as a starting state for the PPO agent and the rest of the episode continues as it would normally (including any exploration that PPO might perform). Only the experiences encountered after the pure exploration phase are on-policy for the PPO agent and therefore only those experiences are used to train it. The experiences collected during the pure exploration phase can be used to optimise a separately trained pure exploration agent (depending on how the pure exploration is implemented). To add some additional stochasticity to the induced starting state distribution, the length of the pure exploration phase is uniformly sampled between 0 and some fixed value $K$ at the start of every episode. 

\section{Experiments}

\subsection{Illustrative CMDP}
\label{sec:exp-ill}
We will first test Explore-Go on the illustrative CMDP from Figure \ref{fig:illustrative}. Training is done on the four tasks in Figure \ref{fig:illustrative-tasks} and unreachable generalisation is evaluated on new tasks with a completely different background colour. For pure exploration, we sample uniformly random actions at each timestep ($\epsilon$-greedy with $\epsilon=1$). We set the maximum length of the pure exploration phase to $K=8$, and compare Explore-Go to a baseline using regular PPO (see Appendix \ref{app:ill} for more details).

In Figure \ref{fig:illustrative-results} we can see that the PPO baseline achieves approximately optimal training performance but is not consistently able to generalise to the unreachable tasks with a different background colour. PPO trains mostly on on-policy data, so when the policy converges to the optimal policy on the training tasks it trains almost exclusively on the on-policy states in Figure \ref{fig:illustrative-optimal}. As we hypothesised before, this likely causes the agent to overfit to the background colour, which will hurt its generalisation capabilities to unreachable states with an unseen background colour. On the other hand, Explore-Go maintains state diversity by performing pure exploration steps at the start of every episode. As such, the state distribution on which it trains resembles the distribution from Figure \ref{fig:illustrative-full}. As we can see in Figure \ref{fig:illustrative-results}, Explore-Go learns slower, but in the end achieves similar training performance to PPO and performs significantly better in the unreachable test tasks. We speculate this is due to the increased diversity of the state distribution on which it trains. 

\begin{figure}[h]
    \centering
    \includegraphics[width=.9\textwidth]{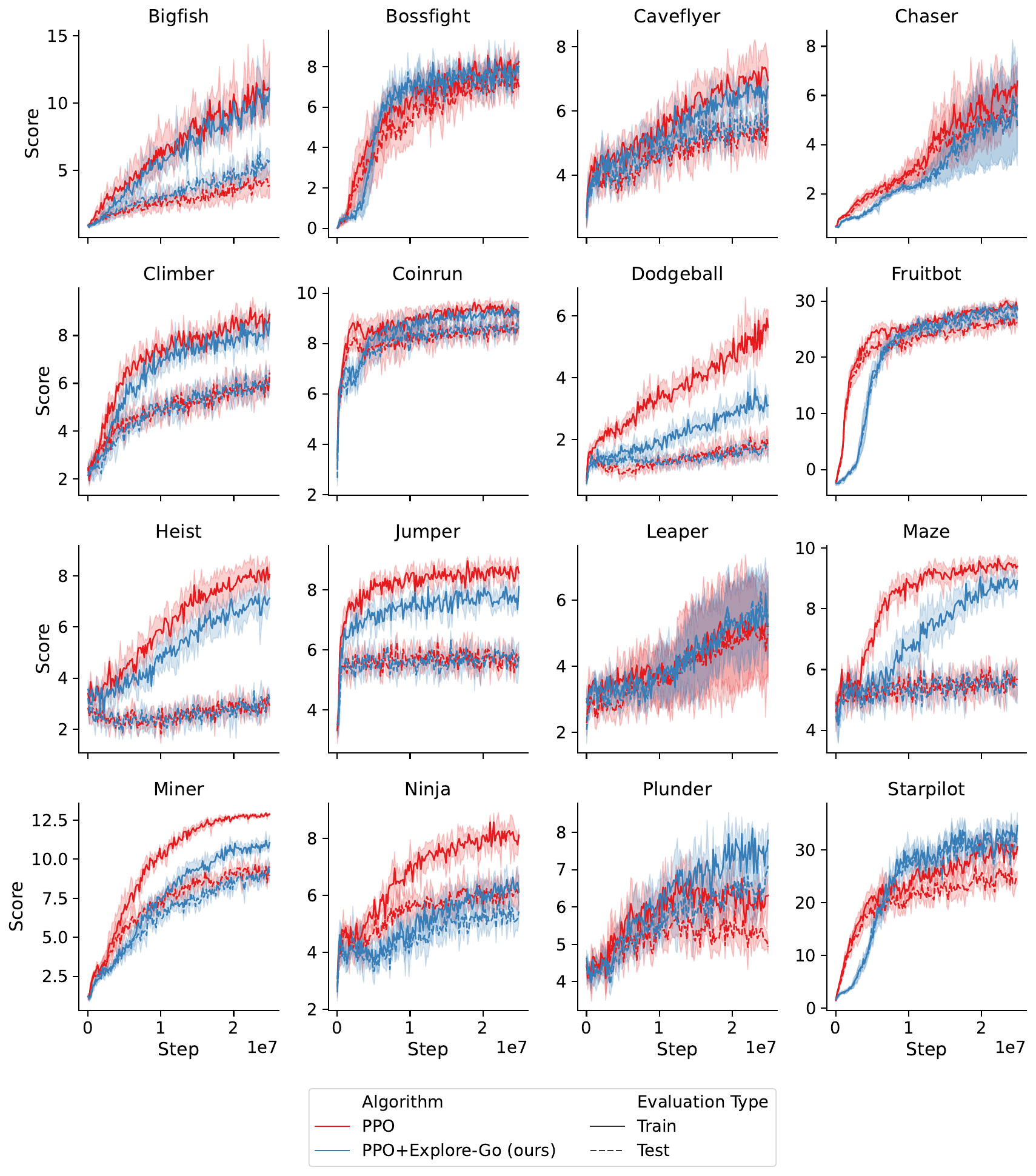}
    \caption{Performance of Explore-Go and PPO on the Procgen Benchmark. Shown are the mean and 95\% confidence interval over 5 seeds.}
    \label{fig:procgen}
\end{figure}

\subsection{Procgen}
\label{sec:exp-proc}
We now test Explore-Go on the popular Procgen benchmark for zero-shot generalisation \citep{cobbe_leveraging_2020}. The Procgen benchmark consists of 16 different environments where every episode a new task (also referred to as \emph{levels} in Procgen) is procedurally generated based on a seed. There are several settings for which the Procgen benchmark can be used but here we use the usual ZSPT setting where the agent gets to train on the first 200 seeds and is tested on 100 seeds randomly sampled from the full task distribution. The task space for the Procgen environments is so large and diverse that we consider all testing tasks unreachable. See Appendix \ref{app:procgen} for more details. 

The pure exploration strategy of sampling random actions we used before is unlikely to yield very diverse states within a reasonable amount of steps. Therefore, we train a separate pure exploration agent with PPO on intrinsic rewards generated through random network distillation \citep[RND,][]{burda_exploration_2019}, a popular deep exploration approach in single-task RL. For Procgen, we set the maximum number of pure exploration steps to $K=200$ for every environment. This number is chosen based on the intuition that the pure exploration phases are long enough for all the environments in Procgen, which have differing average episode lengths. See Appendix \ref{app:procgen} for more experimental details. 

The performance of Explore-Go compared to the baseline PPO is shown in Figure \ref{fig:procgen}. It appears that on 11 out of 16 environments, there is no significant difference in the testing performance of Explore-Go versus PPO. On 2 out of 16 environments (Chaser, Ninja) it looks like Explore-Go might perform slightly worse than PPO. On the remaining 3 environments (Bigfish, Plunder, Starpilot) it appears Explore-Go might achieve better test performance.

Note that the 3 environments in which Explore-Go appears to improve test performance, are all environments in which not moving around much can still lead to significantly different states (due to other objects moving independently from the player). Additionally, in a significant number of the environments, it appears Explore-Go either doesn't have any effect or only results in slowing down the training progress. The slowing down can be explained by the fact that Explore-Go trains on less data than PPO (the steps on the x-axis also include the pure exploration phase at the start of every episode in Explore-Go, which is not used for training the main agent). This leads us to speculate that our pure exploration agent might not be performing very well. This is in line with previous work that has found that exploration approaches used in single-task RL can sometimes perform poorly in procedurally generated environments \citep{raileanu_ride_2020, flet-berliac_adversarially_2021, zhang_noveld_2021, zhang_made_2021, jo_leco_2022, henaff_exploration_2022, henaff_study_2023}.

\section{Related work}
\subsection{Generalisation in CMDPs}
The contextual MDP framework is a very general framework that encompasses many fields in RL that study zero-shot generalisation. For example, the \emph{sim-to-real} setting often encountered in robotics is a special case of the ZSPT setting for CMDPs \citep{kirk_survey_2023}. An approach used to improve generalisation in the sim-to-real setting is domain randomisation \citep{tobin_domain_2017, sadeghi_cad2rl_2017, peng_sim--real_2018}, where the task distribution during training is explicitly increased in order to increase the probability of encompassing the testing tasks in the training distribution. This differs from our work in that we don't explicitly generate more (unreachable) tasks. However, our work could be viewed as implicitly generating more reachable tasks through increased exploration. Another approach that increases the task distribution is data augmentation \citep{raileanu_automatic_2021, lee_network_2020, zhou_domain_2021}. These approaches work by applying a set of given transformations to the states with the prior knowledge that these transformations leave the output (policy or value function) invariant. In this paper, we argue that our approach implicitly induces a form of invariant data augmentation on the states. However, this differs from the other work cited here in that we don't explicitly apply transformations to our states, nor do we require prior knowledge on which transformations leave the policy invariant. 

So far we have mentioned some approaches that increase the number and variability of the training tasks. Other approaches instead try to explicitly bridge the gap between the training and testing tasks. For example, some use inductive biases to encourage learning generalisable functions \citep{zambaldi_relational_2018, zambaldi_deep_2019, kansky_schema_2017, wang_unsupervised_2021, tang_neuroevolution_2020, tang_sensory_2021}. \newpage Others use regularisation techniques from supervised learning to boost generalisation performance \citep{cobbe_quantifying_2019, tishby_deep_2015, igl_generalization_2019, lu_dynamics_2020, eysenbach_robust_2021}. We mention only a selection of approaches here, for a more comprehensive overview we refer to the survey by \citet{kirk_survey_2023}.

All the approaches above use techniques that are not necessarily specific to RL (representation learning, regularisation, etc.). In this work, we instead explore how exploration in RL can be used to improve generalisation.

\subsection{Exploration in CMDPs}
There have been numerous methods of exploration designed specifically for or that have shown promising performance on CMDPs. Some approaches train additional adversarial agents to help with exploration \citep{flet-berliac_adversarially_2021, campero_learning_2021, fickinger_explore_2021}.  Others try to exploit actions that significantly impact the environment \citep{seurin_dont_2021, parisi_interesting_2021} or that cause a significant change in some metric \citep{raileanu_ride_2020, zhang_noveld_2021, zhang_made_2021, ramesh_exploring_2022}. More recently, some approaches have been developed that try to generalise episodic state visitation counts to continuous spaces \citep{jo_leco_2022, henaff_exploration_2022} and several studies have shown the importance of this for exploration in CMDPs \citep{wang_revisiting_2023, henaff_study_2023}. All these methods focus on trading off exploration and exploitation to achieve maximal performance in the training tasks as fast and efficiently as possible. However, in this paper we examine the exploration-exploitation trade-off with respect to maximising generalisation performance in testing tasks. 

In \citet{zisselman_explore_2023}, the authors leverage exploration at test time to move the agent towards states where it can confidently solve the task, thereby increasing test time performance. Our work differs in that we leverage exploration during training time in order to increase the number of states from which the agent can confidently solve the test tasks.  Closest to our work is \citet{weltevrede_role_2023}, \citet{jiang_importance_2023}, \citet{zhu_ingredients_2020} and \citet{suau_bad_2023}. \citet{weltevrede_role_2023} introduce the concepts of reachable generalisation but don't provide intuition on what to do for unreachable generalisation, nor do they propose a novel, scalable approach for maximising generalisation. \citet{jiang_importance_2023} don't make a distinction between reachable and unreachable generalisation and provide intuition which we argue mainly applies to reachable generalisation (see Appendix \ref{app:related}). Moreover, their novel approach only works for off-policy algorithms, whereas ours could be applied to both off-policy and on-policy methods. In \citet{zhu_ingredients_2020}, the authors learn a reset controller that increases the diversity of the agent's start states. However, they only argue (and empirically show) that this benefits reachable generalisation. The concurrent work in \citet{suau_bad_2023} introduces the notion of policy confounding in out-of-trajectory generalisation. The issue of policy confounding is complementary to our intuition for unreachable generalisation. However, it is unclear how out-of-trajectory generalisation equates to reachable or unreachable generalisation. Moreover, they do not propose a novel, scalable approach to solve the issue.

\section{Conclusion}
Recent work has shown that generalisation to reachable \emph{and} unreachable tasks can be improved by exploring and training on more of the reachable state space. They provided intuition for why this happens when generalising to reachable tasks: reachable tasks can be explored and directly optimised during training. However, the intuition for why increasing exploration helps generalising to unreachable tasks was missing. In this work, we provided this intuition by introducing an illustrative CMDP where training on more of the reachable states prevents the agent from overfitting to a spurious correlation between background colour and optimal action. We proposed a novel method Explore-Go based on this intuition that achieves significantly higher test performance than the baseline. Explore-Go effectively increases the starting state distribution for our agent by performing a separate pure exploration phase at the beginning of every episode. Consequently, we postulate the agent trains on significantly more reachable states than it otherwise would and as a result generalises better to unreachable tasks. Finally, we also test our approach on the Procgen benchmark suite and find mixed results. It seems to improve generalisation performance on some environments but not on others. We speculate this is due to an insufficient pure exploration phase which does not result in interesting new starting states.  

One of the limitations of our approach arises if the pure exploration agent is very inefficient. If this is the case, Explore-Go tends to waste interactions with the environment that are not used by the main agent to improve its performance. For future work, we propose to improve the pure exploration agent and investigate ways to increase the starting state distribution of the agent with as little additional environment interactions as possible, for example, by leveraging an off-policy algorithms like DQN \citep{mnih_human-level_2015} or a model-based approach like PEG \citep{hu_planning_2023}.

\section*{Acknowledgements}
This work was partially funded by the Dutch Research Council (NWO) project {\em Reliable Out-of-Distribution Generalization in Deep Reinforcement Learning} with project number OCENW.M.21.234.

\bibliography{bib}

\begin{thebibliography}{61}
\providecommand{\natexlab}[1]{#1}
\providecommand{\url}[1]{\texttt{#1}}
\expandafter\ifx\csname urlstyle\endcsname\relax
  \providecommand{\doi}[1]{doi: #1}\else
  \providecommand{\doi}{doi: \begingroup \urlstyle{rm}\Url}\fi

\bibitem[Akshay et~al.(2013)Akshay, Bertrand, Haddad, and Hélouët]{akshay_steady-state_2013}
S.~Akshay, Nathalie Bertrand, Serge Haddad, and Loïc Hélouët.
\newblock The {Steady}-{State} {Control} {Problem} for {Markov} {Decision} {Processes}.
\newblock In Kaustubh~R. Joshi, Markus Siegle, Mariëlle Stoelinga, and Pedro~R. D'Argenio (eds.), \emph{Quantitative {Evaluation} of {Systems} - 10th {International} {Conference}, {QEST} 2013, {Buenos} {Aires}, {Argentina}, {August} 27-30, 2013. {Proceedings}}, volume 8054 of \emph{Lecture {Notes} in {Computer} {Science}}, pp.\  290--304. Springer, 2013.
\newblock \doi{10.1007/978-3-642-40196-1_26}.
\newblock URL \url{https://doi.org/10.1007/978-3-642-40196-1\_26}.

\bibitem[Bishop(1995)]{bishop_training_1995}
Christopher~M. Bishop.
\newblock Training with {Noise} is {Equivalent} to {Tikhonov} {Regularization}.
\newblock \emph{Neural Comput.}, 7\penalty0 (1):\penalty0 108--116, 1995.
\newblock \doi{10.1162/NECO.1995.7.1.108}.
\newblock URL \url{https://doi.org/10.1162/neco.1995.7.1.108}.

\bibitem[Burda et~al.(2019)Burda, Edwards, Storkey, and Klimov]{burda_exploration_2019}
Yuri Burda, Harrison Edwards, Amos~J. Storkey, and Oleg Klimov.
\newblock Exploration by random network distillation.
\newblock In \emph{7th {International} {Conference} on {Learning} {Representations}, {ICLR} 2019, {New} {Orleans}, {LA}, {USA}, {May} 6-9, 2019}. OpenReview.net, 2019.
\newblock URL \url{https://openreview.net/forum?id=H1lJJnR5Ym}.

\bibitem[Campero et~al.(2021)Campero, Raileanu, Küttler, Tenenbaum, Rocktäschel, and Grefenstette]{campero_learning_2021}
Andres Campero, Roberta Raileanu, Heinrich Küttler, Joshua~B. Tenenbaum, Tim Rocktäschel, and Edward Grefenstette.
\newblock Learning with {AMIGo}: {Adversarially} {Motivated} {Intrinsic} {Goals}.
\newblock In \emph{9th {International} {Conference} on {Learning} {Representations}, {ICLR} 2021, {Virtual} {Event}, {Austria}, {May} 3-7, 2021}. OpenReview.net, 2021.
\newblock URL \url{https://openreview.net/forum?id=ETBc\_MIMgoX}.

\bibitem[Chen et~al.(2020)Chen, Dobriban, and Lee]{chen_group-theoretic_2020}
Shuxiao Chen, Edgar Dobriban, and Jane~H. Lee.
\newblock A {Group}-{Theoretic} {Framework} for {Data} {Augmentation}.
\newblock In Hugo Larochelle, Marc'Aurelio Ranzato, Raia Hadsell, Maria-Florina Balcan, and Hsuan-Tien Lin (eds.), \emph{Advances in {Neural} {Information} {Processing} {Systems} 33: {Annual} {Conference} on {Neural} {Information} {Processing} {Systems} 2020, {NeurIPS} 2020, {December} 6-12, 2020, virtual}, 2020.
\newblock URL \url{https://proceedings.neurips.cc/paper/2020/hash/f4573fc71c731d5c362f0d7860945b88-Abstract.html}.

\bibitem[Chevalier-Boisvert et~al.(2023)Chevalier-Boisvert, Dai, Towers, de~Lazcano, Willems, Lahlou, Pal, Samuel~Castro, and Terry]{chevalier-boisvert_minigrid_2023}
Maxime Chevalier-Boisvert, Bolun Dai, Mark Towers, Rodrigo de~Lazcano, Lucas Willems, Salem Lahlou, Suman Pal, Pablo Samuel~Castro, and Jordan Terry.
\newblock Minigrid {\textbackslash}\& {Miniworld}: {Modular} {\textbackslash}\& {Customizable} {Reinforcement} {Learning} {Environments} for {Goal}-{Oriented} {Tasks}.
\newblock abs/2306.13831, 2023.
\newblock URL \url{https://minigrid.farama.org}.

\bibitem[Cobbe et~al.(2019)Cobbe, Klimov, Hesse, Kim, and Schulman]{cobbe_quantifying_2019}
Karl Cobbe, Oleg Klimov, Christopher Hesse, Taehoon Kim, and John Schulman.
\newblock Quantifying {Generalization} in {Reinforcement} {Learning}.
\newblock In Kamalika Chaudhuri and Ruslan Salakhutdinov (eds.), \emph{Proceedings of the 36th {International} {Conference} on {Machine} {Learning}, {ICML} 2019, 9-15 {June} 2019, {Long} {Beach}, {California}, {USA}}, volume~97 of \emph{Proceedings of {Machine} {Learning} {Research}}, pp.\  1282--1289. PMLR, 2019.
\newblock URL \url{http://proceedings.mlr.press/v97/cobbe19a.html}.

\bibitem[Cobbe et~al.(2020)Cobbe, Hesse, Hilton, and Schulman]{cobbe_leveraging_2020}
Karl Cobbe, Christopher Hesse, Jacob Hilton, and John Schulman.
\newblock Leveraging {Procedural} {Generation} to {Benchmark} {Reinforcement} {Learning}.
\newblock In \emph{Proceedings of the 37th {International} {Conference} on {Machine} {Learning}, {ICML} 2020, 13-18 {July} 2020, {Virtual} {Event}}, volume 119 of \emph{Proceedings of {Machine} {Learning} {Research}}, pp.\  2048--2056. PMLR, 2020.
\newblock URL \url{http://proceedings.mlr.press/v119/cobbe20a.html}.

\bibitem[Cobbe et~al.(2021)Cobbe, Hilton, Klimov, and Schulman]{cobbe_phasic_2021}
Karl Cobbe, Jacob Hilton, Oleg Klimov, and John Schulman.
\newblock Phasic {Policy} {Gradient}.
\newblock In Marina Meila and Tong Zhang (eds.), \emph{Proceedings of the 38th {International} {Conference} on {Machine} {Learning}, {ICML} 2021, 18-24 {July} 2021, {Virtual} {Event}}, volume 139 of \emph{Proceedings of {Machine} {Learning} {Research}}, pp.\  2020--2027. PMLR, 2021.
\newblock URL \url{http://proceedings.mlr.press/v139/cobbe21a.html}.

\bibitem[Ecoffet et~al.(2021)Ecoffet, Huizinga, Lehman, Stanley, and Clune]{ecoffet_first_2021}
Adrien Ecoffet, Joost Huizinga, Joel Lehman, Kenneth~O. Stanley, and Jeff Clune.
\newblock First return, then explore.
\newblock \emph{Nat.}, 590\penalty0 (7847):\penalty0 580--586, 2021.
\newblock \doi{10.1038/S41586-020-03157-9}.
\newblock URL \url{https://doi.org/10.1038/s41586-020-03157-9}.

\bibitem[Espeholt et~al.(2018)Espeholt, Soyer, Munos, Simonyan, Mnih, Ward, Doron, Firoiu, Harley, Dunning, Legg, and Kavukcuoglu]{espeholt_impala_2018}
Lasse Espeholt, Hubert Soyer, Rémi Munos, Karen Simonyan, Volodymyr Mnih, Tom Ward, Yotam Doron, Vlad Firoiu, Tim Harley, Iain Dunning, Shane Legg, and Koray Kavukcuoglu.
\newblock {IMPALA}: {Scalable} {Distributed} {Deep}-{RL} with {Importance} {Weighted} {Actor}-{Learner} {Architectures}.
\newblock In Jennifer~G. Dy and Andreas Krause (eds.), \emph{Proceedings of the 35th {International} {Conference} on {Machine} {Learning}, {ICML} 2018, {Stockholmsmässan}, {Stockholm}, {Sweden}, {July} 10-15, 2018}, volume~80 of \emph{Proceedings of {Machine} {Learning} {Research}}, pp.\  1406--1415. PMLR, 2018.
\newblock URL \url{http://proceedings.mlr.press/v80/espeholt18a.html}.

\bibitem[Eysenbach et~al.(2021)Eysenbach, Salakhutdinov, and Levine]{eysenbach_robust_2021}
Ben Eysenbach, Ruslan Salakhutdinov, and Sergey Levine.
\newblock Robust {Predictable} {Control}.
\newblock In Marc'Aurelio Ranzato, Alina Beygelzimer, Yann~N. Dauphin, Percy Liang, and Jennifer~Wortman Vaughan (eds.), \emph{Advances in {Neural} {Information} {Processing} {Systems} 34: {Annual} {Conference} on {Neural} {Information} {Processing} {Systems} 2021, {NeurIPS} 2021, {December} 6-14, 2021, virtual}, pp.\  27813--27825, 2021.
\newblock URL \url{https://proceedings.neurips.cc/paper/2021/hash/e9f85782949743dcc42079e629332b5f-Abstract.html}.

\bibitem[Feng et~al.(2021)Feng, Gangal, Wei, Chandar, Vosoughi, Mitamura, and Hovy]{feng_survey_2021}
Steven~Y. Feng, Varun Gangal, Jason Wei, Sarath Chandar, Soroush Vosoughi, Teruko Mitamura, and Eduard~H. Hovy.
\newblock A {Survey} of {Data} {Augmentation} {Approaches} for {NLP}.
\newblock In Chengqing Zong, Fei Xia, Wenjie Li, and Roberto Navigli (eds.), \emph{Findings of the {Association} for {Computational} {Linguistics}: {ACL}/{IJCNLP} 2021, {Online} {Event}, {August} 1-6, 2021}, volume ACL/IJCNLP 2021 of \emph{Findings of {ACL}}, pp.\  968--988. Association for Computational Linguistics, 2021.
\newblock \doi{10.18653/V1/2021.FINDINGS-ACL.84}.
\newblock URL \url{https://doi.org/10.18653/v1/2021.findings-acl.84}.

\bibitem[Fickinger et~al.(2021)Fickinger, Jaques, Parajuli, Chang, Rhinehart, Berseth, Russell, and Levine]{fickinger_explore_2021}
Arnaud Fickinger, Natasha Jaques, Samyak Parajuli, Michael Chang, Nicholas Rhinehart, Glen Berseth, Stuart Russell, and Sergey Levine.
\newblock Explore and {Control} with {Adversarial} {Surprise}.
\newblock \emph{CoRR}, abs/2107.07394, 2021.
\newblock URL \url{https://arxiv.org/abs/2107.07394}.
\newblock arXiv: 2107.07394.

\bibitem[Flet-Berliac et~al.(2021)Flet-Berliac, Ferret, Pietquin, Preux, and Geist]{flet-berliac_adversarially_2021}
Yannis Flet-Berliac, Johan Ferret, Olivier Pietquin, Philippe Preux, and Matthieu Geist.
\newblock Adversarially {Guided} {Actor}-{Critic}.
\newblock In \emph{9th {International} {Conference} on {Learning} {Representations}, {ICLR} 2021, {Virtual} {Event}, {Austria}, {May} 3-7, 2021}. OpenReview.net, 2021.
\newblock URL \url{https://openreview.net/forum?id=\_mQp5cr\_iNy}.

\bibitem[Hallak et~al.(2015)Hallak, Castro, and Mannor]{hallak_contextual_2015}
Assaf Hallak, Dotan~Di Castro, and Shie Mannor.
\newblock Contextual {Markov} {Decision} {Processes}.
\newblock \emph{CoRR}, abs/1502.02259, 2015.
\newblock URL \url{http://arxiv.org/abs/1502.02259}.
\newblock arXiv: 1502.02259.

\bibitem[Henaff et~al.(2022)Henaff, Raileanu, Jiang, and Rocktäschel]{henaff_exploration_2022}
Mikael Henaff, Roberta Raileanu, Minqi Jiang, and Tim Rocktäschel.
\newblock Exploration via {Elliptical} {Episodic} {Bonuses}.
\newblock In Sanmi Koyejo, S.~Mohamed, A.~Agarwal, Danielle Belgrave, K.~Cho, and A.~Oh (eds.), \emph{Advances in {Neural} {Information} {Processing} {Systems} 35: {Annual} {Conference} on {Neural} {Information} {Processing} {Systems} 2022, {NeurIPS} 2022, {New} {Orleans}, {LA}, {USA}, {November} 28 - {December} 9, 2022}, 2022.
\newblock URL \url{http://papers.nips.cc/paper\_files/paper/2022/hash/f4f79698d48bdc1a6dec20583724182b-Abstract-Conference.html}.

\bibitem[Henaff et~al.(2023)Henaff, Jiang, and Raileanu]{henaff_study_2023}
Mikael Henaff, Minqi Jiang, and Roberta Raileanu.
\newblock A {Study} of {Global} and {Episodic} {Bonuses} for {Exploration} in {Contextual} {MDPs}.
\newblock In Andreas Krause, Emma Brunskill, Kyunghyun Cho, Barbara Engelhardt, Sivan Sabato, and Jonathan Scarlett (eds.), \emph{International {Conference} on {Machine} {Learning}, {ICML} 2023, 23-29 {July} 2023, {Honolulu}, {Hawaii}, {USA}}, volume 202 of \emph{Proceedings of {Machine} {Learning} {Research}}, pp.\  12972--12999. PMLR, 2023.
\newblock URL \url{https://proceedings.mlr.press/v202/henaff23a.html}.

\bibitem[Hu et~al.(2023)Hu, Chang, Rybkin, and Jayaraman]{hu_planning_2023}
Edward~S. Hu, Richard Chang, Oleh Rybkin, and Dinesh Jayaraman.
\newblock Planning {Goals} for {Exploration}.
\newblock In \emph{The {Eleventh} {International} {Conference} on {Learning} {Representations}, {ICLR} 2023, {Kigali}, {Rwanda}, {May} 1-5, 2023}. OpenReview.net, 2023.
\newblock URL \url{https://openreview.net/forum?id=6qeBuZSo7Pr}.

\bibitem[Igl et~al.(2019)Igl, Ciosek, Li, Tschiatschek, Zhang, Devlin, and Hofmann]{igl_generalization_2019}
Maximilian Igl, Kamil Ciosek, Yingzhen Li, Sebastian Tschiatschek, Cheng Zhang, Sam Devlin, and Katja Hofmann.
\newblock Generalization in {Reinforcement} {Learning} with {Selective} {Noise} {Injection} and {Information} {Bottleneck}.
\newblock In Hanna~M. Wallach, Hugo Larochelle, Alina Beygelzimer, Florence d'Alché Buc, Emily~B. Fox, and Roman Garnett (eds.), \emph{Advances in {Neural} {Information} {Processing} {Systems} 32: {Annual} {Conference} on {Neural} {Information} {Processing} {Systems} 2019, {NeurIPS} 2019, {December} 8-14, 2019, {Vancouver}, {BC}, {Canada}}, pp.\  13956--13968, 2019.
\newblock URL \url{https://proceedings.neurips.cc/paper/2019/hash/e2ccf95a7f2e1878fcafc8376649b6e8-Abstract.html}.

\bibitem[Jiang et~al.(2021)Jiang, Grefenstette, and Rocktäschel]{jiang_prioritized_2021}
Minqi Jiang, Edward Grefenstette, and Tim Rocktäschel.
\newblock Prioritized {Level} {Replay}.
\newblock In Marina Meila and Tong Zhang (eds.), \emph{Proceedings of the 38th {International} {Conference} on {Machine} {Learning}, {ICML} 2021, 18-24 {July} 2021, {Virtual} {Event}}, volume 139 of \emph{Proceedings of {Machine} {Learning} {Research}}, pp.\  4940--4950. PMLR, 2021.
\newblock URL \url{http://proceedings.mlr.press/v139/jiang21b.html}.

\bibitem[Jiang et~al.(2023)Jiang, Kolter, and Raileanu]{jiang_importance_2023}
Yiding Jiang, J.~Zico Kolter, and Roberta Raileanu.
\newblock On the {Importance} of {Exploration} for {Generalization} in {Reinforcement} {Learning}.
\newblock In Alice Oh, Tristan Naumann, Amir Globerson, Kate Saenko, Moritz Hardt, and Sergey Levine (eds.), \emph{Advances in {Neural} {Information} {Processing} {Systems} 36: {Annual} {Conference} on {Neural} {Information} {Processing} {Systems} 2023, {NeurIPS} 2023, {New} {Orleans}, {LA}, {USA}, {December} 10 - 16, 2023}, 2023.
\newblock URL \url{http://papers.nips.cc/paper\_files/paper/2023/hash/2a4310c4fd24bd336aa2f64f93cb5d39-Abstract-Conference.html}.

\bibitem[Jo et~al.(2022)Jo, Kim, Nam, Kwon, Rho, Kim, and Lee]{jo_leco_2022}
DaeJin Jo, Sungwoong Kim, Daniel~Wontae Nam, Taehwan Kwon, Seungeun Rho, Jongmin Kim, and Donghoon Lee.
\newblock {LECO}: {Learnable} {Episodic} {Count} for {Task}-{Specific} {Intrinsic} {Reward}.
\newblock \emph{CoRR}, abs/2210.05409, 2022.
\newblock \doi{10.48550/arXiv.2210.05409}.
\newblock arXiv: 2210.05409.

\bibitem[Kansky et~al.(2017)Kansky, Silver, Mély, Eldawy, Lázaro-Gredilla, Lou, Dorfman, Sidor, Phoenix, and George]{kansky_schema_2017}
Ken Kansky, Tom Silver, David~A. Mély, Mohamed Eldawy, Miguel Lázaro-Gredilla, Xinghua Lou, Nimrod Dorfman, Szymon Sidor, D.~Scott Phoenix, and Dileep George.
\newblock Schema {Networks}: {Zero}-shot {Transfer} with a {Generative} {Causal} {Model} of {Intuitive} {Physics}.
\newblock In Doina Precup and Yee~Whye Teh (eds.), \emph{Proceedings of the 34th {International} {Conference} on {Machine} {Learning}, {ICML} 2017, {Sydney}, {NSW}, {Australia}, 6-11 {August} 2017}, volume~70 of \emph{Proceedings of {Machine} {Learning} {Research}}, pp.\  1809--1818. PMLR, 2017.
\newblock URL \url{http://proceedings.mlr.press/v70/kansky17a.html}.

\bibitem[Kirk et~al.(2023)Kirk, Zhang, Grefenstette, and Rocktäschel]{kirk_survey_2023}
Robert Kirk, Amy Zhang, Edward Grefenstette, and Tim Rocktäschel.
\newblock A {Survey} of {Zero}-shot {Generalisation} in {Deep} {Reinforcement} {Learning}.
\newblock \emph{J. Artif. Intell. Res.}, 76:\penalty0 201--264, 2023.
\newblock \doi{10.1613/JAIR.1.14174}.
\newblock URL \url{https://doi.org/10.1613/jair.1.14174}.

\bibitem[Lee et~al.(2020)Lee, Lee, Shin, and Lee]{lee_network_2020}
Kimin Lee, Kibok Lee, Jinwoo Shin, and Honglak Lee.
\newblock Network {Randomization}: {A} {Simple} {Technique} for {Generalization} in {Deep} {Reinforcement} {Learning}.
\newblock In \emph{8th {International} {Conference} on {Learning} {Representations}, {ICLR} 2020, {Addis} {Ababa}, {Ethiopia}, {April} 26-30, 2020}. OpenReview.net, 2020.
\newblock URL \url{https://openreview.net/forum?id=HJgcvJBFvB}.

\bibitem[Li et~al.(2006)Li, Walsh, and Littman]{li_towards_2006}
Lihong Li, Thomas~J. Walsh, and Michael~L. Littman.
\newblock Towards a {Unified} {Theory} of {State} {Abstraction} for {MDPs}.
\newblock In \emph{International {Symposium} on {Artificial} {Intelligence} and {Mathematics}, {AI}\&{Math} 2006, {Fort} {Lauderdale}, {Florida}, {USA}, {January} 4-6, 2006}, 2006.
\newblock URL \url{http://anytime.cs.umass.edu/aimath06/proceedings/P21.pdf}.

\bibitem[Lin et~al.(2022)Lin, Kaushik, Dyer, and Muthukumar]{lin_good_2022}
Chi-Heng Lin, Chiraag Kaushik, Eva~L. Dyer, and Vidya Muthukumar.
\newblock The good, the bad and the ugly sides of data augmentation: {An} implicit spectral regularization perspective.
\newblock \emph{CoRR}, abs/2210.05021, 2022.
\newblock \doi{10.48550/ARXIV.2210.05021}.
\newblock URL \url{https://doi.org/10.48550/arXiv.2210.05021}.
\newblock arXiv: 2210.05021.

\bibitem[Lu et~al.(2020)Lu, Lee, Abbeel, and Tiomkin]{lu_dynamics_2020}
Xingyu Lu, Kimin Lee, Pieter Abbeel, and Stas Tiomkin.
\newblock Dynamics {Generalization} via {Information} {Bottleneck} in {Deep} {Reinforcement} {Learning}.
\newblock \emph{CoRR}, abs/2008.00614, 2020.
\newblock URL \url{https://arxiv.org/abs/2008.00614}.
\newblock arXiv: 2008.00614.

\bibitem[Lyle et~al.(2020)Lyle, Wilk, Kwiatkowska, Gal, and Bloem-Reddy]{lyle_benefits_2020}
Clare Lyle, Mark van~der Wilk, Marta Kwiatkowska, Yarin Gal, and Benjamin Bloem-Reddy.
\newblock On the {Benefits} of {Invariance} in {Neural} {Networks}.
\newblock \emph{CoRR}, abs/2005.00178, 2020.
\newblock URL \url{https://arxiv.org/abs/2005.00178}.
\newblock arXiv: 2005.00178.

\bibitem[Miao et~al.(2023)Miao, Rainforth, Mathieu, Dubois, Teh, Foster, and Kim]{miao_learning_2023}
Ning Miao, Tom Rainforth, Emile Mathieu, Yann Dubois, Yee~Whye Teh, Adam Foster, and Hyunjik Kim.
\newblock Learning {Instance}-{Specific} {Augmentations} by {Capturing} {Local} {Invariances}.
\newblock In Andreas Krause, Emma Brunskill, Kyunghyun Cho, Barbara Engelhardt, Sivan Sabato, and Jonathan Scarlett (eds.), \emph{International {Conference} on {Machine} {Learning}, {ICML} 2023, 23-29 {July} 2023, {Honolulu}, {Hawaii}, {USA}}, volume 202 of \emph{Proceedings of {Machine} {Learning} {Research}}, pp.\  24720--24736. PMLR, 2023.
\newblock URL \url{https://proceedings.mlr.press/v202/miao23a.html}.

\bibitem[Mnih et~al.(2015)Mnih, Kavukcuoglu, Silver, Rusu, Veness, Bellemare, Graves, Riedmiller, Fidjeland, Ostrovski, Petersen, Beattie, Sadik, Antonoglou, King, Kumaran, Wierstra, Legg, and Hassabis]{mnih_human-level_2015}
Volodymyr Mnih, Koray Kavukcuoglu, David Silver, Andrei~A. Rusu, Joel Veness, Marc~G. Bellemare, Alex Graves, Martin~A. Riedmiller, Andreas Fidjeland, Georg Ostrovski, Stig Petersen, Charles Beattie, Amir Sadik, Ioannis Antonoglou, Helen King, Dharshan Kumaran, Daan Wierstra, Shane Legg, and Demis Hassabis.
\newblock Human-level control through deep reinforcement learning.
\newblock \emph{Nat.}, 518\penalty0 (7540):\penalty0 529--533, 2015.
\newblock \doi{10.1038/NATURE14236}.
\newblock URL \url{https://doi.org/10.1038/nature14236}.

\bibitem[Moon et~al.(2022)Moon, Lee, and Song]{moon_rethinking_2022}
Seungyong Moon, JunYeong Lee, and Hyun~Oh Song.
\newblock Rethinking {Value} {Function} {Learning} for {Generalization} in {Reinforcement} {Learning}.
\newblock In Sanmi Koyejo, S.~Mohamed, A.~Agarwal, Danielle Belgrave, K.~Cho, and A.~Oh (eds.), \emph{Advances in {Neural} {Information} {Processing} {Systems} 35: {Annual} {Conference} on {Neural} {Information} {Processing} {Systems} 2022, {NeurIPS} 2022, {New} {Orleans}, {LA}, {USA}, {November} 28 - {December} 9, 2022}, 2022.
\newblock URL \url{http://papers.nips.cc/paper\_files/paper/2022/hash/e19ab2dde2e60cf68d1ded18c38938f4-Abstract-Conference.html}.

\bibitem[Parisi et~al.(2021)Parisi, Dean, Pathak, and Gupta]{parisi_interesting_2021}
Simone Parisi, Victoria Dean, Deepak Pathak, and Abhinav Gupta.
\newblock Interesting {Object}, {Curious} {Agent}: {Learning} {Task}-{Agnostic} {Exploration}.
\newblock In Marc'Aurelio Ranzato, Alina Beygelzimer, Yann~N. Dauphin, Percy Liang, and Jennifer~Wortman Vaughan (eds.), \emph{Advances in {Neural} {Information} {Processing} {Systems} 34: {Annual} {Conference} on {Neural} {Information} {Processing} {Systems} 2021, {NeurIPS} 2021, {December} 6-14, 2021, virtual}, pp.\  20516--20530, 2021.
\newblock URL \url{https://proceedings.neurips.cc/paper/2021/hash/abe8e03e3ac71c2ec3bfb0de042638d8-Abstract.html}.

\bibitem[Peng et~al.(2018)Peng, Andrychowicz, Zaremba, and Abbeel]{peng_sim--real_2018}
Xue~Bin Peng, Marcin Andrychowicz, Wojciech Zaremba, and Pieter Abbeel.
\newblock Sim-to-{Real} {Transfer} of {Robotic} {Control} with {Dynamics} {Randomization}.
\newblock In \emph{2018 {IEEE} {International} {Conference} on {Robotics} and {Automation}, {ICRA} 2018, {Brisbane}, {Australia}, {May} 21-25, 2018}, pp.\  1--8. IEEE, 2018.
\newblock \doi{10.1109/ICRA.2018.8460528}.

\bibitem[Raileanu \& Fergus(2021)Raileanu and Fergus]{raileanu_decoupling_2021}
Roberta Raileanu and Rob Fergus.
\newblock Decoupling {Value} and {Policy} for {Generalization} in {Reinforcement} {Learning}.
\newblock In Marina Meila and Tong Zhang (eds.), \emph{Proceedings of the 38th {International} {Conference} on {Machine} {Learning}, {ICML} 2021, 18-24 {July} 2021, {Virtual} {Event}}, volume 139 of \emph{Proceedings of {Machine} {Learning} {Research}}, pp.\  8787--8798. PMLR, 2021.
\newblock URL \url{http://proceedings.mlr.press/v139/raileanu21a.html}.

\bibitem[Raileanu \& Rocktäschel(2020)Raileanu and Rocktäschel]{raileanu_ride_2020}
Roberta Raileanu and Tim Rocktäschel.
\newblock {RIDE}: {Rewarding} {Impact}-{Driven} {Exploration} for {Procedurally}-{Generated} {Environments}.
\newblock In \emph{8th {International} {Conference} on {Learning} {Representations}, {ICLR} 2020, {Addis} {Ababa}, {Ethiopia}, {April} 26-30, 2020}. OpenReview.net, 2020.
\newblock URL \url{https://openreview.net/forum?id=rkg-TJBFPB}.

\bibitem[Raileanu et~al.(2021)Raileanu, Goldstein, Yarats, Kostrikov, and Fergus]{raileanu_automatic_2021}
Roberta Raileanu, Max Goldstein, Denis Yarats, Ilya Kostrikov, and Rob Fergus.
\newblock Automatic {Data} {Augmentation} for {Generalization} in {Reinforcement} {Learning}.
\newblock In Marc'Aurelio Ranzato, Alina Beygelzimer, Yann~N. Dauphin, Percy Liang, and Jennifer~Wortman Vaughan (eds.), \emph{Advances in {Neural} {Information} {Processing} {Systems} 34: {Annual} {Conference} on {Neural} {Information} {Processing} {Systems} 2021, {NeurIPS} 2021, {December} 6-14, 2021, virtual}, pp.\  5402--5415, 2021.
\newblock URL \url{https://proceedings.neurips.cc/paper/2021/hash/2b38c2df6a49b97f706ec9148ce48d86-Abstract.html}.

\bibitem[Ramesh et~al.(2022)Ramesh, Kirsch, Steenkiste, and Schmidhuber]{ramesh_exploring_2022}
Aditya Ramesh, Louis Kirsch, Sjoerd~van Steenkiste, and Jürgen Schmidhuber.
\newblock Exploring through {Random} {Curiosity} with {General} {Value} {Functions}.
\newblock In Sanmi Koyejo, S.~Mohamed, A.~Agarwal, Danielle Belgrave, K.~Cho, and A.~Oh (eds.), \emph{Advances in {Neural} {Information} {Processing} {Systems} 35: {Annual} {Conference} on {Neural} {Information} {Processing} {Systems} 2022, {NeurIPS} 2022, {New} {Orleans}, {LA}, {USA}, {November} 28 - {December} 9, 2022}, 2022.
\newblock URL \url{http://papers.nips.cc/paper\_files/paper/2022/hash/76e57c3c6b3e06f332a4832ddd6a9a12-Abstract-Conference.html}.

\bibitem[Sadeghi \& Levine(2017)Sadeghi and Levine]{sadeghi_cad2rl_2017}
Fereshteh Sadeghi and Sergey Levine.
\newblock {CAD2RL}: {Real} {Single}-{Image} {Flight} {Without} a {Single} {Real} {Image}.
\newblock In Nancy~M. Amato, Siddhartha~S. Srinivasa, Nora Ayanian, and Scott Kuindersma (eds.), \emph{Robotics: {Science} and {Systems} {XIII}, {Massachusetts} {Institute} of {Technology}, {Cambridge}, {Massachusetts}, {USA}, {July} 12-16, 2017}, 2017.
\newblock \doi{10.15607/RSS.2017.XIII.034}.
\newblock URL \url{http://www.roboticsproceedings.org/rss13/p34.html}.

\bibitem[Schulman et~al.(2017)Schulman, Wolski, Dhariwal, Radford, and Klimov]{schulman_proximal_2017}
John Schulman, Filip Wolski, Prafulla Dhariwal, Alec Radford, and Oleg Klimov.
\newblock Proximal {Policy} {Optimization} {Algorithms}.
\newblock \emph{CoRR}, abs/1707.06347, 2017.
\newblock URL \url{http://arxiv.org/abs/1707.06347}.
\newblock arXiv: 1707.06347.

\bibitem[Seurin et~al.(2021)Seurin, Strub, Preux, and Pietquin]{seurin_dont_2021}
Mathieu Seurin, Florian Strub, Philippe Preux, and Olivier Pietquin.
\newblock Don't {Do} {What} {Doesn}'t {Matter}: {Intrinsic} {Motivation} with {Action} {Usefulness}.
\newblock In Zhi-Hua Zhou (ed.), \emph{Proceedings of the {Thirtieth} {International} {Joint} {Conference} on {Artificial} {Intelligence}, {IJCAI} 2021, {Virtual} {Event} / {Montreal}, {Canada}, 19-27 {August} 2021}, pp.\  2950--2956. ijcai.org, 2021.
\newblock \doi{10.24963/IJCAI.2021/406}.
\newblock URL \url{https://doi.org/10.24963/ijcai.2021/406}.

\bibitem[Shen et~al.(2022)Shen, Bubeck, and Gunasekar]{shen_data_2022}
Ruoqi Shen, Sébastien Bubeck, and Suriya Gunasekar.
\newblock Data {Augmentation} as {Feature} {Manipulation}.
\newblock In Kamalika Chaudhuri, Stefanie Jegelka, Le~Song, Csaba Szepesvári, Gang Niu, and Sivan Sabato (eds.), \emph{International {Conference} on {Machine} {Learning}, {ICML} 2022, 17-23 {July} 2022, {Baltimore}, {Maryland}, {USA}}, volume 162 of \emph{Proceedings of {Machine} {Learning} {Research}}, pp.\  19773--19808. PMLR, 2022.
\newblock URL \url{https://proceedings.mlr.press/v162/shen22a.html}.

\bibitem[Shorten \& Khoshgoftaar(2019)Shorten and Khoshgoftaar]{shorten_survey_2019}
Connor Shorten and Taghi~M. Khoshgoftaar.
\newblock A survey on {Image} {Data} {Augmentation} for {Deep} {Learning}.
\newblock \emph{J. Big Data}, 6:\penalty0 60, 2019.
\newblock \doi{10.1186/S40537-019-0197-0}.
\newblock URL \url{https://doi.org/10.1186/s40537-019-0197-0}.

\bibitem[Suau et~al.(2023)Suau, Spaan, and Oliehoek]{suau_bad_2023}
Miguel Suau, Matthijs T.~J. Spaan, and Frans~A. Oliehoek.
\newblock Bad {Habits}: {Policy} {Confounding} and {Out}-of-{Trajectory} {Generalization} in {RL}.
\newblock \emph{CoRR}, abs/2306.02419, 2023.
\newblock \doi{10.48550/ARXIV.2306.02419}.
\newblock URL \url{https://doi.org/10.48550/arXiv.2306.02419}.
\newblock arXiv: 2306.02419.

\bibitem[Sutton \& Barto(2018)Sutton and Barto]{sutton_reinforcement_2018}
Richard~S. Sutton and Andrew~G. Barto.
\newblock \emph{Reinforcement learning: an introduction}.
\newblock Adaptive computation and machine learning series. The MIT Press, Cambridge, Massachusetts, second edition edition, 2018.
\newblock ISBN 978-0-262-03924-6.

\bibitem[Tang \& Ha(2021)Tang and Ha]{tang_sensory_2021}
Yujin Tang and David Ha.
\newblock The {Sensory} {Neuron} as a {Transformer}: {Permutation}-{Invariant} {Neural} {Networks} for {Reinforcement} {Learning}.
\newblock In Marc'Aurelio Ranzato, Alina Beygelzimer, Yann~N. Dauphin, Percy Liang, and Jennifer~Wortman Vaughan (eds.), \emph{Advances in {Neural} {Information} {Processing} {Systems} 34: {Annual} {Conference} on {Neural} {Information} {Processing} {Systems} 2021, {NeurIPS} 2021, {December} 6-14, 2021, virtual}, pp.\  22574--22587, 2021.
\newblock URL \url{https://proceedings.neurips.cc/paper/2021/hash/be3e9d3f7d70537357c67bb3f4086846-Abstract.html}.

\bibitem[Tang et~al.(2020)Tang, Nguyen, and Ha]{tang_neuroevolution_2020}
Yujin Tang, Duong Nguyen, and David Ha.
\newblock Neuroevolution of self-interpretable agents.
\newblock In Carlos Artemio~Coello Coello (ed.), \emph{{GECCO} '20: {Genetic} and {Evolutionary} {Computation} {Conference}, {Cancún} {Mexico}, {July} 8-12, 2020}, pp.\  414--424. ACM, 2020.
\newblock \doi{10.1145/3377930.3389847}.

\bibitem[Tishby \& Zaslavsky(2015)Tishby and Zaslavsky]{tishby_deep_2015}
Naftali Tishby and Noga Zaslavsky.
\newblock Deep learning and the information bottleneck principle.
\newblock In \emph{2015 {IEEE} {Information} {Theory} {Workshop}, {ITW} 2015, {Jerusalem}, {Israel}, {April} 26 - {May} 1, 2015}, pp.\  1--5. IEEE, 2015.
\newblock \doi{10.1109/ITW.2015.7133169}.

\bibitem[Tobin et~al.(2017)Tobin, Fong, Ray, Schneider, Zaremba, and Abbeel]{tobin_domain_2017}
Josh Tobin, Rachel Fong, Alex Ray, Jonas Schneider, Wojciech Zaremba, and Pieter Abbeel.
\newblock Domain randomization for transferring deep neural networks from simulation to the real world.
\newblock In \emph{2017 {IEEE}/{RSJ} {International} {Conference} on {Intelligent} {Robots} and {Systems}, {IROS} 2017, {Vancouver}, {BC}, {Canada}, {September} 24-28, 2017}, pp.\  23--30. IEEE, 2017.
\newblock \doi{10.1109/IROS.2017.8202133}.

\bibitem[Wang et~al.(2023)Wang, Zhou, Kang, Feng, and Yan]{wang_revisiting_2023}
Kaixin Wang, Kuangqi Zhou, Bingyi Kang, Jiashi Feng, and Shuicheng Yan.
\newblock Revisiting {Intrinsic} {Reward} for {Exploration} in {Procedurally} {Generated} {Environments}.
\newblock In \emph{The {Eleventh} {International} {Conference} on {Learning} {Representations}, {ICLR} 2023, {Kigali}, {Rwanda}, {May} 1-5, 2023}. OpenReview.net, 2023.
\newblock URL \url{https://openreview.net/pdf?id=j3GK3\_xZydY}.

\bibitem[Wang et~al.(2021)Wang, Lian, and Yu]{wang_unsupervised_2021}
Xudong Wang, Long Lian, and Stella~X. Yu.
\newblock Unsupervised {Visual} {Attention} and {Invariance} for {Reinforcement} {Learning}.
\newblock In \emph{{IEEE} {Conference} on {Computer} {Vision} and {Pattern} {Recognition}, {CVPR} 2021, virtual, {June} 19-25, 2021}, pp.\  6677--6687. Computer Vision Foundation / IEEE, 2021.
\newblock \doi{10.1109/CVPR46437.2021.00661}.
\newblock URL \url{https://openaccess.thecvf.com/content/CVPR2021/html/Wang\_Unsupervised\_Visual\_Attention\_and\_Invariance\_for\_Reinforcement\_Learning\_CVPR\_2021\_paper.html}.

\bibitem[Weltevrede et~al.(2023)Weltevrede, Spaan, and Böhmer]{weltevrede_role_2023}
Max Weltevrede, Matthijs T.~J. Spaan, and Wendelin Böhmer.
\newblock The {Role} of {Diverse} {Replay} for {Generalisation} in {Reinforcement} {Learning}.
\newblock \emph{CoRR}, abs/2306.05727, 2023.
\newblock \doi{10.48550/ARXIV.2306.05727}.
\newblock URL \url{https://doi.org/10.48550/arXiv.2306.05727}.
\newblock arXiv: 2306.05727.

\bibitem[Zambaldi et~al.(2018)Zambaldi, Raposo, Santoro, Bapst, Li, Babuschkin, Tuyls, Reichert, Lillicrap, Lockhart, Shanahan, Langston, Pascanu, Botvinick, Vinyals, and Battaglia]{zambaldi_relational_2018}
Vinícius~Flores Zambaldi, David Raposo, Adam Santoro, Victor Bapst, Yujia Li, Igor Babuschkin, Karl Tuyls, David~P. Reichert, Timothy~P. Lillicrap, Edward Lockhart, Murray Shanahan, Victoria Langston, Razvan Pascanu, Matthew~M. Botvinick, Oriol Vinyals, and Peter~W. Battaglia.
\newblock Relational {Deep} {Reinforcement} {Learning}.
\newblock \emph{CoRR}, abs/1806.01830, 2018.
\newblock URL \url{http://arxiv.org/abs/1806.01830}.
\newblock arXiv: 1806.01830.

\bibitem[Zambaldi et~al.(2019)Zambaldi, Raposo, Santoro, Bapst, Li, Babuschkin, Tuyls, Reichert, Lillicrap, Lockhart, Shanahan, Langston, Pascanu, Botvinick, Vinyals, and Battaglia]{zambaldi_deep_2019}
Vinícius~Flores Zambaldi, David Raposo, Adam Santoro, Victor Bapst, Yujia Li, Igor Babuschkin, Karl Tuyls, David~P. Reichert, Timothy~P. Lillicrap, Edward Lockhart, Murray Shanahan, Victoria Langston, Razvan Pascanu, Matthew~M. Botvinick, Oriol Vinyals, and Peter~W. Battaglia.
\newblock Deep reinforcement learning with relational inductive biases.
\newblock In \emph{7th {International} {Conference} on {Learning} {Representations}, {ICLR} 2019, {New} {Orleans}, {LA}, {USA}, {May} 6-9, 2019}. OpenReview.net, 2019.
\newblock URL \url{https://openreview.net/forum?id=HkxaFoC9KQ}.

\bibitem[Zhang et~al.(2021{\natexlab{a}})Zhang, Bengio, Hardt, Recht, and Vinyals]{zhang_understanding_2021}
Chiyuan Zhang, Samy Bengio, Moritz Hardt, Benjamin Recht, and Oriol Vinyals.
\newblock Understanding deep learning (still) requires rethinking generalization.
\newblock \emph{Commun. ACM}, 64\penalty0 (3):\penalty0 107--115, 2021{\natexlab{a}}.
\newblock \doi{10.1145/3446776}.

\bibitem[Zhang et~al.(2021{\natexlab{b}})Zhang, Rashidinejad, Jiao, Tian, Gonzalez, and Russell]{zhang_made_2021}
Tianjun Zhang, Paria Rashidinejad, Jiantao Jiao, Yuandong Tian, Joseph~E Gonzalez, and Stuart Russell.
\newblock {MADE}: {Exploration} via {Maximizing} {Deviation} from {Explored} {Regions}.
\newblock In \emph{Advances in {Neural} {Information} {Processing} {Systems}}, volume~34, pp.\  9663--9680. Curran Associates, Inc., 2021{\natexlab{b}}.
\newblock URL \url{https://proceedings.neurips.cc/paper/2021/hash/5011bf6d8a37692913fce3a15a51f070-Abstract.html}.

\bibitem[Zhang et~al.(2021{\natexlab{c}})Zhang, Xu, Wang, Wu, Keutzer, Gonzalez, and Tian]{zhang_noveld_2021}
Tianjun Zhang, Huazhe Xu, Xiaolong Wang, Yi~Wu, Kurt Keutzer, Joseph~E. Gonzalez, and Yuandong Tian.
\newblock {NovelD}: {A} {Simple} yet {Effective} {Exploration} {Criterion}.
\newblock In Marc'Aurelio Ranzato, Alina Beygelzimer, Yann~N. Dauphin, Percy Liang, and Jennifer~Wortman Vaughan (eds.), \emph{Advances in {Neural} {Information} {Processing} {Systems} 34: {Annual} {Conference} on {Neural} {Information} {Processing} {Systems} 2021, {NeurIPS} 2021, {December} 6-14, 2021, virtual}, pp.\  25217--25230, 2021{\natexlab{c}}.
\newblock URL \url{https://proceedings.neurips.cc/paper/2021/hash/d428d070622e0f4363fceae11f4a3576-Abstract.html}.

\bibitem[Zhou et~al.(2021)Zhou, Yang, Qiao, and Xiang]{zhou_domain_2021}
Kaiyang Zhou, Yongxin Yang, Yu~Qiao, and Tao Xiang.
\newblock Domain {Generalization} with {MixStyle}.
\newblock In \emph{9th {International} {Conference} on {Learning} {Representations}, {ICLR} 2021, {Virtual} {Event}, {Austria}, {May} 3-7, 2021}. OpenReview.net, 2021.
\newblock URL \url{https://openreview.net/forum?id=6xHJ37MVxxp}.

\bibitem[Zhu et~al.(2020)Zhu, Yu, Gupta, Shah, Hartikainen, Singh, Kumar, and Levine]{zhu_ingredients_2020}
Henry Zhu, Justin Yu, Abhishek Gupta, Dhruv Shah, Kristian Hartikainen, Avi Singh, Vikash Kumar, and Sergey Levine.
\newblock The {Ingredients} of {Real} {World} {Robotic} {Reinforcement} {Learning}.
\newblock In \emph{8th {International} {Conference} on {Learning} {Representations}, {ICLR} 2020, {Addis} {Ababa}, {Ethiopia}, {April} 26-30, 2020}. OpenReview.net, 2020.
\newblock URL \url{https://openreview.net/forum?id=rJe2syrtvS}.

\bibitem[Zisselman et~al.(2023)Zisselman, Lavie, Soudry, and Tamar]{zisselman_explore_2023}
Ev~Zisselman, Itai Lavie, Daniel Soudry, and Aviv Tamar.
\newblock Explore to {Generalize} in {Zero}-{Shot} {RL}.
\newblock In Alice Oh, Tristan Naumann, Amir Globerson, Kate Saenko, Moritz Hardt, and Sergey Levine (eds.), \emph{Advances in {Neural} {Information} {Processing} {Systems} 36: {Annual} {Conference} on {Neural} {Information} {Processing} {Systems} 2023, {NeurIPS} 2023, {New} {Orleans}, {LA}, {USA}, {December} 10 - 16, 2023}, 2023.
\newblock URL \url{http://papers.nips.cc/paper\_files/paper/2023/hash/c793577b644268259b1416464a6cdb8c-Abstract-Conference.html}.

\end{thebibliography}
\bibliographystyle{bibstyle}

\newpage

\appendix

\section{Discussion on related work}
\label{app:related}
\citet{jiang_importance_2023} argue that generalisation in RL extends beyond representation learning. They do so with an example in a tabular grid-world environment. In the environment they describe the agent during training always starts in the top left corner of the grid, and the goal is always in the top right corner. During testing the agent starts in a different position in the grid-world (in their example, the lower left corner). This is according to our definition an example of a reachable task. They then argue (in the way we described in Section \ref{sec:background-expl}) that more exploration can improve generalisation to these tasks. 

They extend their intuition to non-tabular CMDPs by arguing that in certain cases two states that are unreachable from each other, can nonetheless inside a neural network map to similar representations. As a result, even though a state in the input space is unreachable, it can be mapped to something reachable in the latent representational space and therefore the reachable generalisation arguments apply again. For this reason, the generalisation benefits from more exploration can go beyond representation learning. 

Relating it to the illustrative example we provide in Figure \ref{fig:illustrative}, we argue this intuition considers the generalisation benefits one might obtain from learning to act optimally in more abstracted states. For example, in \citet{jiang_importance_2023}'s grid-world the lower states would have normally unseen values, which is represented by increasing the number of columns on which we train in Figure \ref{fig:illustrative-optimal} and \ref{fig:illustrative-full}. However, in Section \ref{sec:unreachable} we argue that specifically unreachable generalisation can benefit as well from training on more states belonging to the same abstracted states (represented by increasing the number of rows on which we train in Figure \ref{fig:illustrative-optimal} and \ref{fig:illustrative-full}). Training on more of these states could encourage the agent to learn representations that map different unreachable states to the same latent representation (or equivalently, abstracted states). As such, we argue the generalisation benefits from more exploration can in part be attributed to an implicit form of representation learning (which some experiments performed by \citet{weltevrede_role_2023} seem to corroborate). 

\section{Experimental details}
\label{app:exp-details}

\subsection{Illustrative CMDP}
\label{app:ill}
The training tasks for the illustrative CMDP experiment in Section \ref{sec:exp-ill} are the ones depicted in Figure~\ref{fig:illustrative-tasks}. The unreachable testing tasks consist of 4 tasks with the same starting positions as found in the training tasks (the end-point of the arms) but with a white background colour. The states the agent observes are structured as RGB images with shape $(3, 5, 5)$. The entire $5 \times 5$ grid is encoded with the background colour of the particular task, except for the goal position (at $(2,2)$) which is dark green ((0,0.5,0) in RGB) and the agent (wherever it is located at that time) which is dark red ((0.5,0,0) in RGB).  The specific background colours are the following:
\begin{itemize}
    \item \textbf{Training task 1:} (0,0,1)
    \item \textbf{Training task 2:} (0,1,0)
    \item \textbf{Training task 3:} (1,0,0)
    \item \textbf{Training task 4:} (1,0,1)
    \item \textbf{Testing tasks:} (1,1,1)
\end{itemize}

Moving into a wall of the cross will leave the agent position unchanged, except for the additional transitions between the cross endpoints. Moving into the goal position (middle of the cross) will terminate the episode and give a reward of 1. All other transitions give a reward of 0. The agent is timed out after 20 steps. 

\begin{table}[h]
\centering
\begin{tabular}{@{}ll@{}}
\toprule
\textbf{Hyper-parameter}               & \textbf{Value}     \\ \midrule
Total timesteps                       & 50 000            \\
Vectorised environments               & 4                 \\ \midrule
\multicolumn{2}{c}{\textbf{PPO}}                           \\
timesteps per rollout               & 10           \\
epochs per rollout                  & 3             \\
minibatches per epoch               & 8                \\
Discount factor $\gamma$              & 0.9               \\
GAE smoothing parameter ($\lambda$)   & 0.95                \\
Entropy bonus                         & 0.01                \\
PPO clip range ($\epsilon$)           & 0.2                 \\
Reward normalisation?                 & No                  \\
Max. gradient norm                    & .5                  \\
Shared actor and critic networks      & No                 \\   \midrule
\multicolumn{2}{c}{\textbf{Adam}}                          \\
Learning rate                         & $1 \times 10^{-4}$ \\
Epsilon                               & $1 \times 10^{-5}$ \\           
\end{tabular}
\caption{Hyper-parameters used for the illustrative CMDP experiment}
\label{tab:par-ill}
\end{table}

\subsubsection*{Implementation details}
For PPO we used the implementation by \citet{moon_rethinking_2022} which we adapted for PPO + Explore-Go. The hyperparameters for both PPO and PPO + Explore-Go can be found in Table \ref{tab:par-ill}. The only additional hyperparameter that Explore-Go uses is the maximal number of pure exploration steps $K$, which we choose to be $K=8$. Both algorithms use network architectures that flatten the $(3, 5, 5)$ observation and feed it through a fully connected network with a ReLU activation function. The hidden dimensions for both the actor and critic are $[128, 64, 32]$ followed by an output layer of size $[1]$ for the critic and size $[|A|]$ for the actor. The output of the actor is used as logits in a categorical distribution over the actions.

\subsection{Procgen}
\label{app:procgen}
The Procgen benchmark consists of 16 environments that are inspired by classic video games. The observations are $64 \times 64$ RGB images and there are a total of 15 actions. Not all actions are operational in all environments. For example, in the Bigfish environment the agent can only move along in an 8-directional space (up, down, left, right and the diagonals), the other actions don't do anything in this environment. 

At the start of every episode, the level for that episode is procedurally generated based on a seed. Variations between levels can include things like the background image, the topology of the environment and/or the number, type or movement of obstacles/enemies. For more details we refer to \citet{cobbe_leveraging_2020}. 

\begin{table}[h]
\centering
\begin{tabular}{@{}ll@{}}
\toprule
\textbf{Hyper-parameter}               & \textbf{Value}     \\ \midrule
Total timesteps                       & $25 \times 10^6$            \\
Vectorised environments               & 64                 \\ \midrule
\multicolumn{2}{c}{\textbf{PPO}}                           \\
timesteps per rollout               & 256           \\
epochs per rollout                  & 3             \\
minibatches per epoch               & 8                \\
Discount factor $\gamma$              & 0.999               \\
GAE smoothing parameter ($\lambda$)   & 0.95                \\
Entropy bonus                         & 0.01                \\
PPO clip range ($\epsilon$)           & 0.2                 \\
Reward normalisation?                 & Yes                  \\
Max. gradient norm                    & .5                  \\
Shared actor and critic networks      & Yes                 \\   \midrule
\multicolumn{2}{c}{\textbf{Adam}}                          \\
Learning rate                         & $5 \times 10^{-4}$ \\
Epsilon                               & $1 \times 10^{-5}$ \\           
\end{tabular}
\caption{Hyper-parameters used for the Procgen experiment}
\label{tab:par-proc}
\end{table}

\subsubsection*{Implementation details}
For PPO we used the implementation by \citet{moon_rethinking_2022} which we adapted for PPO + Explore-Go. The hyperparameters for both PPO and PPO + Explore-Go can be found in Table \ref{tab:par-proc} and are the same as in \citet{cobbe_leveraging_2020}. The pure exploration agent in Explore-Go trained on the intrinsic rewards from an RND network is trained with PPO with the same hyperparameters. The additional hyperparameters for the pure exploration agent can be found in Table \ref{tab:par-pure}. Note that the number of transitions collected per rollout is fixed (256*64). However, in the Explore-Go method this includes both pure exploration and the normal agent phase. This means that both the pure exploration agent and the regular agent are trained on a variable number of transitions after each rollout (different from the PPO agent that always trains on the fixed 256*64 transitions). Both algorithms use critic and actor networks using ResNet architectures from \citet{espeholt_impala_2018} as was also done in previous work \citep{cobbe_leveraging_2020, cobbe_phasic_2021, moon_rethinking_2022}.  

The RND predictor network is trained on the observations encountered during the pure exploration phase and the intrinsic reward for a transition $\langle s,a,s'\rangle$ is defined as the squared $L_2$ norm of the difference between the output of the predictor and target network evaluated on the next state $s'$. The intrinsic rewards are normalised by dividing by the running estimate of the standard deviation. The RND target and predictor networks have the same ResNet architecture as the actor and critic used by the agent but differ in that the last hidden layer of size 256 used by the critic and actor (mapped to a single value or distribution over actions in the critic and actor respectively) is replaced by two hidden layers of size 1024 which are finally mapped to an embedding dimension (output layer) of size 512.

\begin{table}[h]
\centering
\begin{tabular}{@{}ll@{}}
\toprule
\textbf{Hyper-parameter}               & \textbf{Value}     \\ \midrule
Max number of pure exploration steps ($K$)     & 200                 \\ \midrule
\multicolumn{2}{c}{\textbf{RND}}                           \\
Learning rate               & $1 \times 10^{-4}$           \\         
Embedding dimension         & 512                           \\
\end{tabular}
\caption{Hyper-parameters used for the Procgen experiment}
\label{tab:par-pure}
\end{table}

\end{document}